\documentclass[runningheads,oneside]{llncs}

\usepackage{PRIMEarxiv}

\usepackage[utf8]{inputenc} 
\usepackage[T1]{fontenc}    
\usepackage{hyperref}       
\usepackage{url}            
\usepackage{booktabs}       
\usepackage{amsfonts}       
\usepackage{nicefrac}       
\usepackage{microtype}      
\usepackage{lipsum}
\usepackage{fancyhdr}       
\usepackage{graphicx}       
\graphicspath{{media/}}     
\usepackage{caption}
\usepackage{float}
\usepackage{xcolor}
\usepackage{hyperref}
\usepackage{authblk}        
\usepackage{adjustbox}
\usepackage{lineno}         
\usepackage{listings}
\usepackage{fvextra}
\usepackage{amsmath}
\usepackage{authblk}

\usepackage{multirow}

\definecolor{orange}{HTML}{CC8B00}
\definecolor{green}{HTML}{596318}
\definecolor{red}{HTML}{FD151B}

\title{Specialized curricula for training vision-language models in retinal image analysis}

\author[1]{Robbie Holland}
\author[2]{Thomas R. P. Taylor}
\author[3]{Christopher Holmes}
\author[4]{Sophie Riedl}
\author[4]{Julia Mai}
\author[2]{Maria Patsiamanidi}
\author[2]{Dimitra Mitsopoulou}
\author[5]{Paul Hager}
\author[5]{Philip Müller}
\author[1,5]{Johannes C. Paetzold}
\author[6,7,8]{Hendrik P. N. Scholl}
\author[4]{Hrvoje Bogunović}
\author[4]{Ursula Schmidt-Erfurth}
\author[1,5,$\dagger$]{Daniel Rueckert}
\author[3,$\dagger$]{Sobha Sivaprasad}
\author[2,$\dagger$]{Andrew J. Lotery}
\author[1,5,$\dagger$]{and Martin J. Menten}
\author[ ]{on behalf of the PINNACLE consortium}

\affil[1]{Biomedical Image Analysis, Department of Computing, Imperial College London, United Kingdom}
\affil[2]{Clinical and Experimental Sciences, Faculty of Medicine, University of Southampton, United Kingdom}
\affil[3]{Moorfields Eye Hospital NHS Foundation Trust, London, United Kingdom}
\affil[4]{Laboratory for Ophthalmic Image Analysis, Medical University of Vienna, Austria}
\affil[5]{Institute for AI in Healthcare and Medicine, Klinikum rechts der Isar, Technical University of Munich, Germany}
\affil[6]{Institute of Molecular and Clinical Ophthalmology Basel, Switzerland}
\affil[7]{Department of Ophthalmology, University of Basel, Switzerland}
\affil[8]{Department of Clinical Pharmacology, Medical University of Vienna, Austria}

\begin{document}
\maketitle

\begingroup
\let\thefootnote\relax\footnotetext{$^{\dagger}$ denotes shared senior authorship}
\endgroup

\begin{abstract}

Clinicians spend a significant amount of time reviewing medical images and transcribing their findings regarding patient diagnosis, referral and treatment in text form. Vision-language models (VLMs), which automatically interpret images and summarize their findings as text, have enormous potential to alleviate clinical workloads and increase patient access to high-quality medical care. While foundational models have stirred considerable interest in the medical community, it is unclear whether their general capabilities translate to real-world clinical utility. In this work, we demonstrate that {{} OpenAI's ChatGPT-4o model, in addition to two foundation VLMs designed for medical use,} markedly underperform compared to practicing ophthalmologists on specialist tasks crucial to the care of patients with age-related macular degeneration (AMD). To address this, we initially identified the essential capabilities required for image-based clinical decision-making, and then developed a curriculum to selectively train VLMs in these skills. The resulting model, RetinaVLM, can be instructed to write reports that significantly outperform those written by leading foundation medical VLMs {{} and ChatGPT-4o} in disease staging (F1 score of 0.63 vs. {{} 0.33}) and patient referral (0.67 vs. {{} 0.50}), and approaches the diagnostic performance of junior ophthalmologists (who achieve 0.77 and 0.78 on the respective tasks). Furthermore, in a single-blind reader study two senior ophthalmologists with up to 32 years of experience found RetinaVLM's reports were found to be {{} substantially more accurate than those by ChatGPT-4o (64.3\% vs. 14.3\%)}. These results reinforce that our curriculum-based approach provides a blueprint towards specializing foundation medical VLMs for real-world clinical tasks.
\end{abstract}

\section{Introduction}

\begin{figure}
    \centering
    \includegraphics[width=0.92\linewidth]{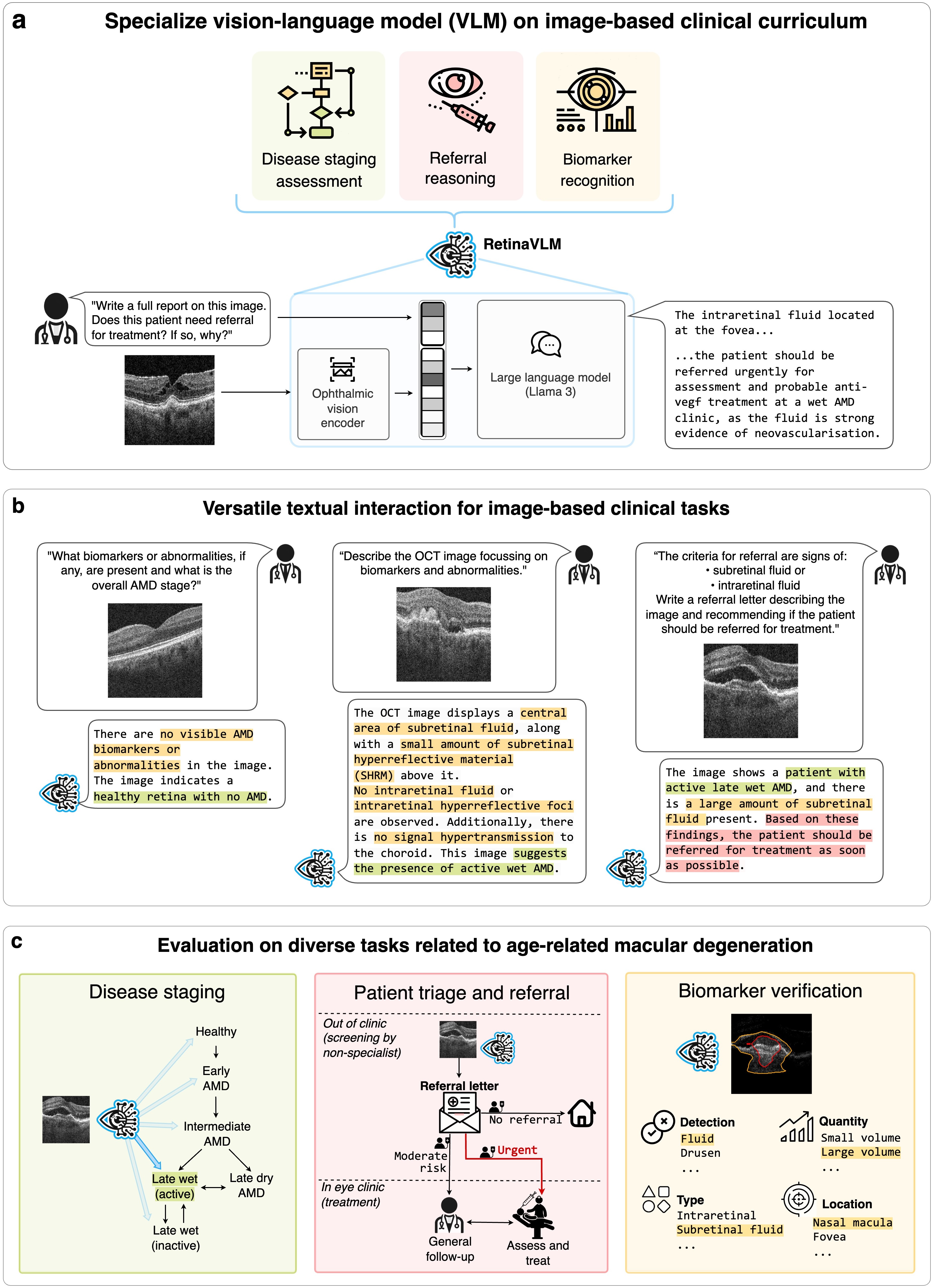}
    \caption{{{} We introduce RetinaVLM, a specialist medical generative vision-language model (VLM).} (\textbf{a}) Using a {{} curriculum-based approach}, we trained RetinaVLM in specialist medical skills that medical foundation VLMs are currently lacking (\textbf{b}) RetinaVLM is able to process retinal optical retinal optical coherence tomography (OCT) images and flexibly respond to text-based queries. (\textbf{c}) Its abilities entail the analysis of imaging biomarkers of age-related macular degeneration (AMD), disease staging, and the referral for treatment.}
    \label{fig:study_overview}
\end{figure}

Medical images are central to many clinical decisions regarding patient diagnosis, referral, and treatment. Clinicians spend a significant amount of time transcribing image-based decisions into text in order to store and communicate their findings \cite{moy2021measurement,acosta2022multimodal}. Visual-language models (VLM), which automatically interpret medical images and generate detailed textual descriptions, have enormous potential to alleviate clinical workloads and increase patient access to high-quality medical care \cite{moor2023foundation,rajpurkar2023current}. To date, the majority of medical VLMs have been trained to output a finite set of pre-determined textual responses \cite{zhang2022contrastive,huang2023visual,lu2024visual,christensen2024vision}. Only recently, the combination of large language models (LLM) with medical vision encoders has led to the development of more powerful and versatile \textit{generative} VLMs that are able to write comprehensive text reports or answer complex questions \cite{li2023llava,moor2023med,tu2024towards}.

This current generation of medical language models is fueled by vast amounts of unstructured training data that is extracted from medical textbooks, scientific publications or social media posts of healthcare professionals \cite{li2023llava,huang2023visual,moor2023med}. These \textit{foundation} language models have stirred considerable interest among the medical community for their expert-level performance on standardized medical question-answering tasks, such as licensing exams and case studies \cite{kung2023performance,singhal2023large}. However, it is unclear whether this general performance translates to clinical utility in specialist medical domains \cite{hager2024evaluating}. { Despite its impressive scale, the training data of foundation language models has been collected agnostically of their downstream application. The resulting model is unlikely to acquire the nuanced knowledge necessary for effective application in specialized clinical contexts.}


In this study, we identify this missing piece in foundation models towards developing generative medical VLMs with real-world clinical utility. We propose to deconstruct clinical problems into sets of mandatory capabilities required for their resolution and selectively train VLMs in these skills. 
{{} To train VLMs in these skills, we develop a curriculum-based approach \cite{bengio2009curriculum} that draws from recent advances in instruction finetuning \cite{ouyang2022training,wei2021finetuned,liu2024visual} which iteratively refine models on datasets of increasing quality.}
{ We demonstrate the feasibility of this approach in ophthalmology, focussing our analysis to a single retinal disease enables us to assess the clinical limitations of foundation VLMs, and the benefits of training on specialist curricula, in depth.}
{{} To this end we introduce, RetinaVLM, is a generative medical VLM for OCT images (see Figure \ref{fig:study_overview}a).} RetinaVLM is trained using a {{} two-part} dedicated curriculum that is specific to the clinical management of age-related macular degeneration (AMD), the leading cause of blindness in the elderly \cite{mitchell2018age,wong2014global}. The resulting model is able to process optical coherence tomography (OCT) images of the retina and flexibly respond to instructions and questions (see Figure \ref{fig:study_overview}b). In particular, we evaluate RetinaVLM's utility and versatility regarding disease staging, patient referral and biomarker analysis in AMD (see Figure \ref{fig:study_overview}c).

\section{Results}
\subsection{RetinaVLM, a specialist vision-language model for retinal image analysis}

RetinaVLM combines two main components: an ophthalmic vision encoder that processes input OCT images, and a generative LLM that handles textual instructions and outputs the corresponding responses (see Figure \ref{fig:study_overview}a). The vision encoder {was trained using self-supervised learning on images from the train set, and was found to perform on par with RETFound \cite{holland2024metadata}, a large foundation model for retinal image analysis \cite{zhou2023foundation}. For the language model, we use Meta's Llama 3 as generative LLM which was the most performant, openly available model at the time of this study \cite{MetaLlama3}. Both these deep neural networks have already been pre-trained on large OCT and natural language datasets, respectively. {{} We combine these to create RetinaVLM, following the architectural design of MiniGPT-4 introduced by Zhu et al. \cite{zhu2023minigpt}.} This approach leaves the pretrained models unchanged, and instead trains a third, intermediary network to map visual information from the image encoder to the language model. Additional information regarding the model architecture, training and inference are detailed in Section \ref{method:vlm_model}.
}

{
\subsection{A curriculum to encode the capabilities of retinal specialists into vision-language models}
}
\label{sec:curriculum}

{{}
An intuitive strategy to specialize VLMs while preserving their ability to flexibly interact with text queries is to provide them with a set of medical images and corresponding \textit{visual question-answer} (VQA) pairs. VQA-based training, a form of instruction fine-tuning \cite{anderson2018bottom,liu2024visual}, guides VLMs to extract visual information  and synthesize it into textual outputs that adhere to specific user directives. This approach also introduces diversity through varying instructions and responses, mitigating overfitting while enhancing the model's capacity to selectively report relevant clinical features and contextualize findings. However, relevant VQA datasets are scarce for most medical specializations, including ophthalmology.
}


Together with a large team of ophthalmologists, which are involved with the patient care and academic research of AMD, we {{} created a curriculum of VQA datasets designed to train VLMs for assisting image-based clinical decisions regarding AMD. To this end, the ophthalmologists first} defined a set of guidelines outlining essential capabilities of agents assisting the image-based clinical management of AMD {{}(verbose versions are documented in Figure \ref{extended_fig:curricula})}. Specifically, these include details relating to the identification of AMD biomarkers in OCT images, the linking of these to the AMD disease stage, and ultimately deciding on the required referral and treatment of the patient. These subsequently guided {{} a combination of manual and automated efforts to} curate a training curriculum, which consists of 41,926 OCT images, and 479,710 VQA pairs to progressively specialize VLMs in these capabilities.

\begin{figure}
    \centering
    \includegraphics[width=0.95\linewidth]{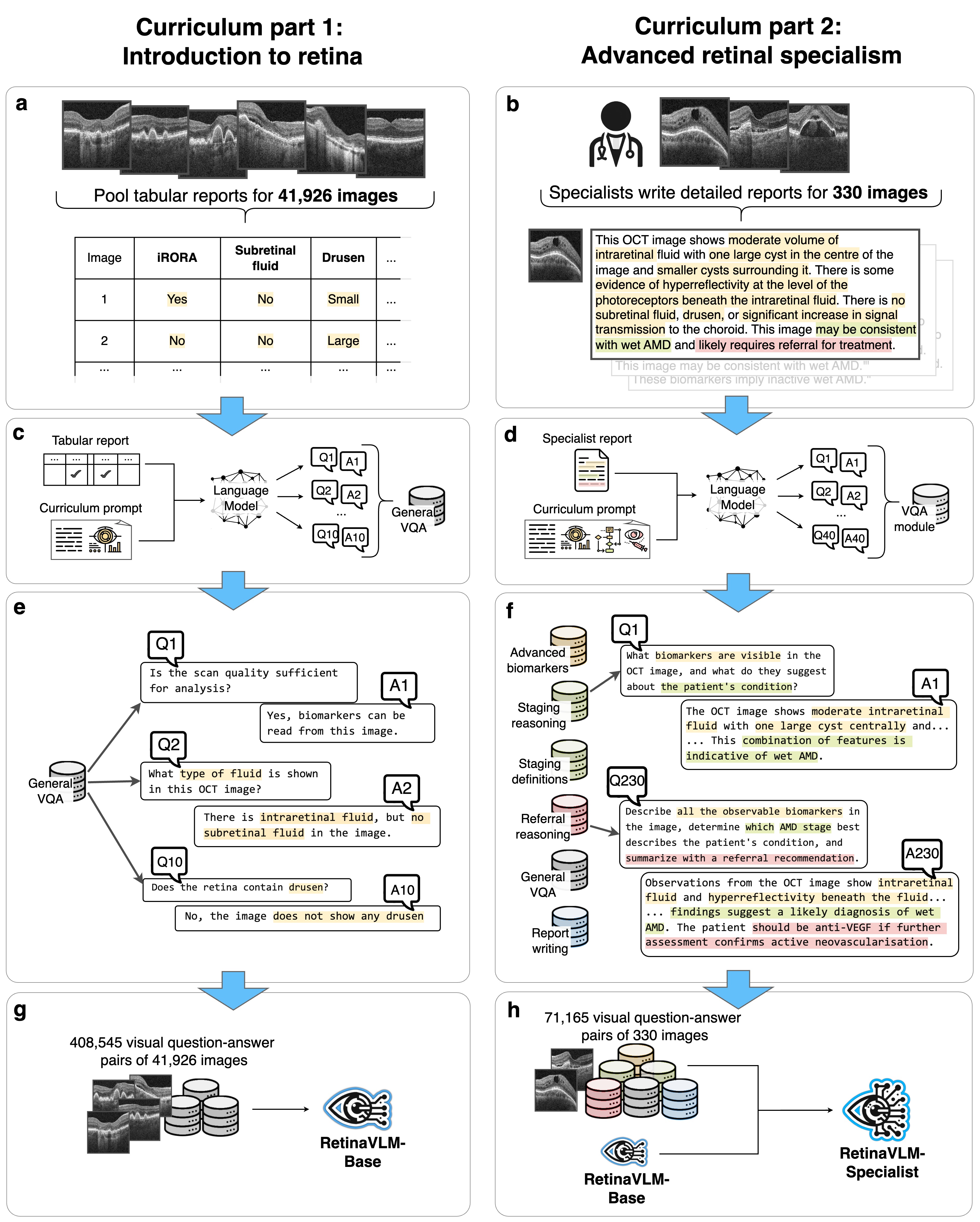}
    \caption{We curated a two-part curriculum to specialize medical VLMs for clinical use. (\textbf{a} and \textbf{b}) Based on a retrospectively collected OCT imaging dataset, we created a large number of tabular reports as well as a small number of comprehensive textual reports. (\textbf{c} and \textbf{d}) We then used an independent LLM to automatically generate visual question-answers based on these reports. (\textbf{e} and \textbf{f}) This yielded two VQA datasets, the first on basic imaging biomarkers of AMD and the second covering more advanced clinical skills. (\textbf{g} and \textbf{h}) Finally, we trained two specialist medical generative VLMs, RetinaVLM-Base and RetinaVLM-Specialist, using either the first or both VQA datasets.}
    \label{fig:figure_2}
\end{figure}

\paragraph{Curriculum part 1: Introduction to retina}
\label{sec:intro_to_retina}

The first part of the curriculum, named \textit{Introduction to retina}, primarily covers the appearance of the retina and AMD biomarkers in OCT images. Using automated data collection, we obtained tabular reports for 41,926 retrospectively collected OCT images of AMD patients (see Figure \ref{fig:figure_2}a). Each report describes the visible biomarkers, patient's diagnosis, visual acuity and demographic information in 34 data fields. A full description of the OCT dataset can be found in Section \ref{sec:method:oct_dataset}, the list of all tabular data fields and example tabular reports in Figure \ref{extended_fig:tabular_reports}, and the methodology for their automated procurement in Section \ref{sec:method:tabular}.

Next, we tasked an independent LLM to generate question-answer pairs based on these reports (see Figure \ref{fig:figure_2}c). The model processed the content of the tabular reports -- but not the OCT images -- to output a numbered list of question-answer pairs. We generated an average of ten question-answer pairs per report that are mostly related to the presence or absence of specific biomarkers (see Figure \ref{fig:figure_2}e). The LLM was instructed to create both closed-ended 'yes or no' style questions, and simple open-ended questions. Detailed information on the LLM setup can be found in Section \ref{sec:method:tabular}.

This automated approach allowed us to generate a large dataset of 408,545 question-answer pairs. However, the questions were limited in scope to the set of biomarkers documented by the tabular reports. Training on these yielded the first of two specialist VLMs, \textit{RetinaVLM-Base} (see Figure \ref{fig:figure_2}g).

\paragraph{Curriculum part 2: Advanced retinal specialism}
\label{sec:advanced_specialism}

The second part of the curriculum, named \textit{Advanced retinal specialism}, builds on top of the first part to link imaging biomarkers to AMD stage and the recommended course of treatment. As this reasoning cannot be fully conveyed via tabular information, we tasked two ophthalmologists with 3 and 10 years of experience, respectively, to create comprehensive textual reports for a subset of 330 OCT images (see Figure \ref{fig:figure_2}b). The ophthalmologists were asked to primarily describe the main pathological biomarkers related to AMD while also noting any other observations regarding the retinal anatomy. This task yielded high-quality reports that go beyond the short notes that are typically written by ophthalmologists in their clinical routine. Instructions given to the ophthalmologists as well as a set of sample reports are provided in Section \ref{sec:method:specialist} and Figure \ref{extended_fig:specialist_reports}, respectively.

Similar to before, an independent LLM was then employed to automatically generate question-answer pairs based on the reports (see Figure \ref{fig:figure_2}d). Due to the substantially increased depth and scope of the full-text reports compared to the tabular ones, we used several advanced LLM instructions to create 216 diverse question-answer pairs per image on average (see Figure \ref{fig:figure_2}f). These cover additional biomarkers and sub-categorize them based on their size, type, and location. Other question-answer pairs are related to the causal relationship between biomarkers and six AMD disease stages as well as three levels of patient referral urgency. Moreover, the question-answer pairs were more varied in their structure in order to preserve interactive capabilities of the foundation LLM. For example, some queries asked to summarize the existing reports or provide several answers in succession. An example interaction with the LLM to generate question-answer pairs with the LLM is shown in Figure \ref{extended_fig:qa_generation}. Furthermore, a list of all the LLM instructions is provided in Section \ref{supplemental:qa_prompts} and example question-answers yielded by this approach are shown in the `part 2' section of Figure \ref{extended_fig:qa_examples}.

This resulted in a dataset of 71,165 advanced question-answer pairs. By further training RetinaVLM-Base on the second part of the curriculum, we obtained our most performant VLM for the clinical management of AMD, \textit{RetinaVLM-Specialist} (see Figure \ref{fig:figure_2}h).

\subsection{RetinaVLM-Specialist outperforms foundation models and approaches junior ophthalmologists in AMD disease staging and report writing}
\label{sec:results_staging}

\begin{figure}
    \centering
    \includegraphics[width=0.98\linewidth]{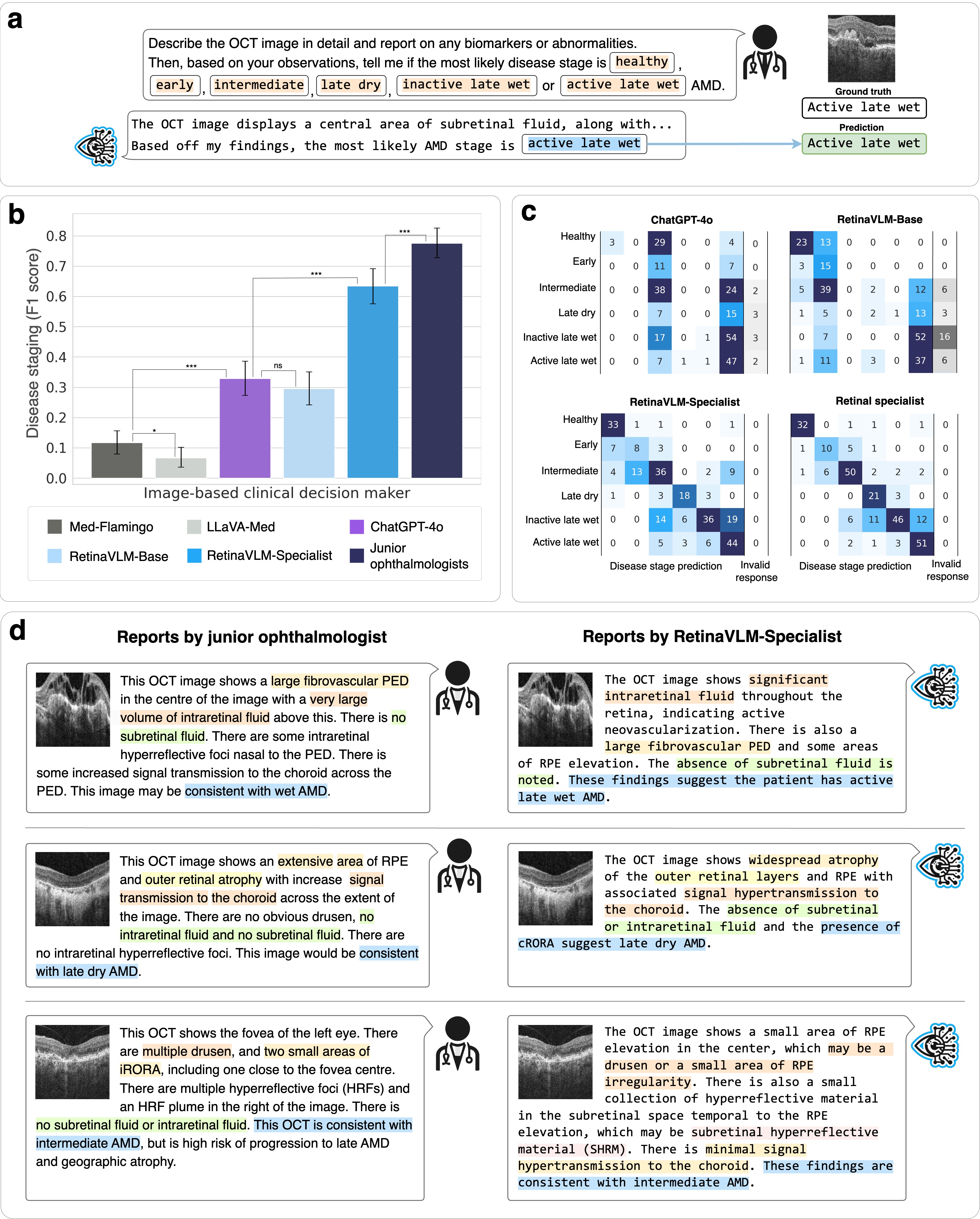}
    \caption{(\textbf{a}) Comparison of the ability of four VLMs to write reports on retinal OCT images and derive the AMD stage. (\textbf{b}) Overall staging accuracy for each model was calculated using micro F1 scores with 95\% CI, with tests of statistical significance calculated using McNemar's test. (\textbf{c}) Confusion matrices between the senior ophthalmologists' assessments (rows) against the image-based clinical decision maker's prediction (columns). (\textbf{d}) Qualitative comparison of reports written by human ophthalmologists and RetinaVLM-Specialist with text markings highlighting findings regarding biomarker observations and disease stage.}
    \label{fig:figure_3}
\end{figure}

{{} AMD is a debilitating and irreversible condition marked by the progressive loss of central vision, severely hindering essential activities like reading, driving, and recognizing faces. The demand for timely diagnosis and management currently exceeds the available ophthalmology expertise \cite{resnikoff2012number}. Moreover, largely due to aging populations, projections indicate a nearly 50\% increase in AMD cases globally to nearly 300 million by 2040 \cite{wong2014global}. In this context, automated retinal image analysis emerges as an essential tool to support the interpretation and textual reporting of retinal images.

A key aspect of image report generation involves estimating the disease stage indicated by the retinal image. We assessed the ability of five} different generative VLMs to determine the AMD disease stage {{} when writing reports on} retinal OCT images. Specifically, we benchmarked {{} two foundation VLMs, Med-Flamingo \cite{moor2023med} and LLaVA-Med \cite{li2023llava}, with purported general abilities in medical image analysis. In addition, we assessed OpenAI's ChatGPT-4o model \cite{openai2024gpt4o}, developed by OpenAI, L.P. (San Francisco, California, USA). We compared these baseline models against RetinaVLM-Base and RetinaVLM-Specialist, which result from cumulative training on curriculum part 1 and part 2, respectively. Finally we compared these automated models against the overall performance of the six junior ophthalmologists. This experiment was conducted using a testing dataset} of 276 previously unseen OCT images, on which VLMs were tasked to write descriptive reports before classifying the patient into one of six disease stages (see Figure \ref{fig:figure_3}a). The model predictions were compared to ground truth labels obtained from ophthalmologists. { Each image was initially graded by two out of six junior ophthalmologists, {{} who have  2, 3, 5, 8, 10 and 15 years of experience working full-time in ophthalmology clinics after receiving their medical degree}. Inter-rater disagreements were resolved by a panel of two senior ophthalmologists with 25 and 32 years of experience, respectively (see Section \ref{supplemental:junior_senior_ophthalmologists} for additional details and roles of the junior and senior ophthalmologists).} For additional methodological details {{} relating to the the derivation of testing labels and } the instruction given to all VLMs to generate these reports, see Sections \ref{sec:method:oct_dataset} and \ref{sec:method:disease_staging}.

{{} We found that both the intermediate RetinaVLM-Base model and ChatGPT-4o perform significantly better than Med-Flamingo and LLaVA-Med, which lack the ophthalmological specialism to stage disease (see Figure \ref{fig:figure_3}b). By more effectively classifying conversion to late stages of AMD RetinaVLM-Base and ChatGPT-4o achieve F1 scores of 0.33 and 0.29, respectively. However, beyond this distinction, these models failed to differentiate finer disease stage variations and were} markedly outperformed by RetinaVLM-Specialist which scored 0.63 F1. This performance approached, but did not match, the accuracy of the junior ophthalmologists who achieved an F1 score of 0.78. We analyze this last discrepancy in further detail in Section \ref{sec:discussion_limitations}. Moreover, foundation VLMs and RetinaVLM-Base returned a substantial number of invalid reports that did not conclude with one of the six disease stages (see Figure \ref{fig:figure_3}c). Conversely, all generated reports by RetinaVLM-Specialist were valid. Similar to human experts, RetinaVLM-Specialist struggled the most when diagnosing wet inactive AMD. We attribute this to the high number of shared imaging biomarkers that indicate either intermediate and late-wet forms of AMD, which sometimes leads to misdiagnosis by both ophthalmologists and RetinaVLM-Specialist (see Figure \ref{fig:figure_3}d). Four additional examples of success and failure cases of RetinaVLM-Specialist are shown in Figure \ref{extended_fig:extra_success_failure}a. Moreover, full numerical results, including a comparison to a standard image-only classification model, as well as the confusion matrix for Med-Flamingo, are shown in Figure \ref{extended_fig:results_tables}a and \ref{extended_fig:med-flamingo_confusion}, respectively.

\begin{figure}
    \centering
    \includegraphics[width=0.98\linewidth]{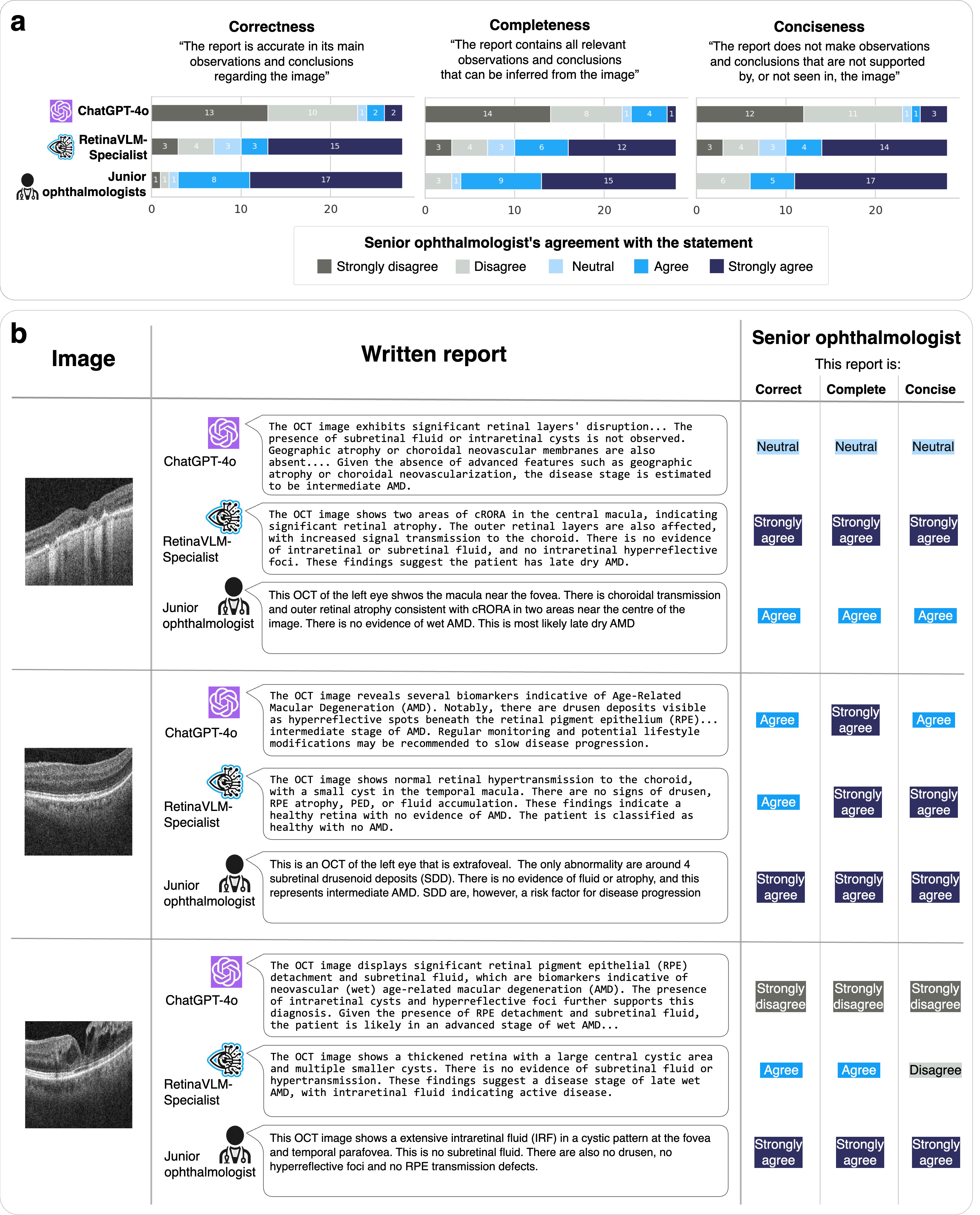}
    \caption{(\textbf{a}) {{} Summary statistics of the quality of image reports written by ChatGPT-4o, RetinaVLM-Specialist and junior ophthalmologists, broken down by correctness, completeness and conciseness.} Reports were scored for on each of the three criteria by senior ophthalmologists using a five-point Likert scale. (\textbf{b}) Representative reports with ratings by one of the senior ophthalmologists. {{} As ChatGPT-4o tended to write excessively long reports, despite being prompted to shorten them, we display passages the senior ophthalmologists selected as the most important to their given rating. For verbose versions and additional sample reports by ChatGPT-4o see Supplementary Figure \ref{fig:chatgpt_output}.}}
    \label{fig:figure_4}
\end{figure}

{{}
\subsubsection{Direct evaluation of imaging reports by senior ophthalmologists}
}
\label{sec:results_ccc_evaluation}
Next, 84 of the generated reports were scored by the two senior ophthalmologists for their correctness, completeness, and conciseness. They were shown 28 reports written by {{} ChatGPT-4o}, 28 by RetinaVLM-Specialist, and 28 by the two annotating junior ophthalmologists. {{} We then randomly ordered the 84 reports to decrease the likelihood that multiple reports regarding the same image would appear in succession. Moreover, to mitigate bias we conducted this evaluation as a single-blind study in which the authorship of each report was concealed from the senior ophthalmologists.} For each report, the senior ophthalmologist first reviewed the corresponding OCT image before rating the generated report in the three criteria on a five point Likert scale \cite{van2024adapted}.

{ The senior ophthalmologists observed that {{} ChatGPT-4o largely} failed to compose factually correct image reports (see Figure \ref{fig:figure_4}a), even though it uses specialist terminology that may give the reports the initial appearance of being written by an ophthalmologist (see Figure \ref{fig:figure_4}b).} {{} ChatGPT-4o was found to consistently hallucinate the presence of subretinal fluid, and in every relevant instance failed to detect presence of subretinal hyperreflective material and hypertransmission, which are crucial for diagnosing inactive late wet and late dry AMD, respectively. We have included six sample reports by ChatGPT-4o which contain examples of these errors in Supplementary Figure \ref{fig:chatgpt_output}, including verbose versions of the three displayed in Figure \ref{fig:figure_4}b. Overall, the senior ophthalmologists found that only 4 out of 28 (14.3\%) of the reports written by ChatGPT-4o were correct in their observations and conclusions.}

{{} Overall senior ophthalmologists either agreed or strongly agreed that reports generated by RetinaVLM-Specialist were more correct (18 vs. 4, or 64.3\% vs. 14.3\%), complete (18 vs. 5) and concise (18 vs. 4) than those written by ChatGPT-4o. Compared to ChatGPT-4o, RetinaVLM-Specialist correctly detected a wider range of biomarkers, which more frequently led to a correct disease stage estimation. However, there remains a gap in overall performance between RetinaVLM-Specialist and the junior ophthalmologists, with the senior ophthalmologists rating their reports to be more correct (25 vs. 18), complete (24 vs. 18) and concise (22 vs. 18).

An example of this discrepancy can be seen in} the second sample in  \ref{fig:figure_4}b, where RetinaVLM-Specialist correctly identified that the image showed a healthy retina, but also detects a small cyst that was not found in the image. Junior ophthalmologists wrote a concise report, but incorrectly associated subretinal drusenoid deposits with intermediate AMD. {{} Moreover, both RetinaVLM, which has currently only been trained to identify AMD from retinal imaging biomarkers, and  ChatGPT-4o fail to report likely cases of retinal vascular disease.} These characteristics of both the junior ophthalmologists and RetinaVLM-Specialist are discussed in more detail in Section \ref{sec:discussion_limitations}.

\subsection{RetinaVLM-Specialist surpasses opticians and approaches junior ophthalmologists in AMD patient screening and referral}
\label{sec:results_referral}

\begin{figure}
    \centering
    \includegraphics[width=0.92\linewidth]{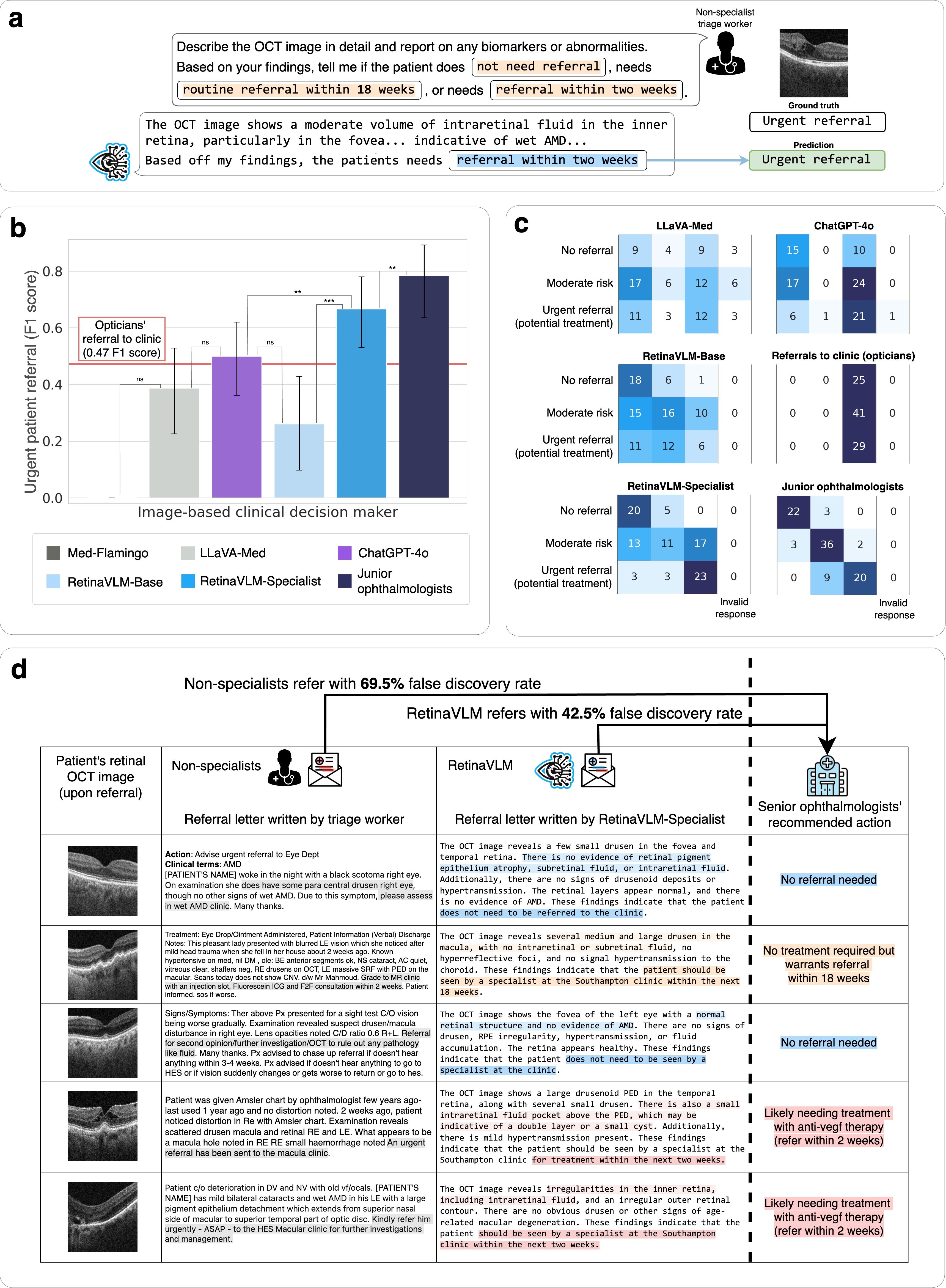}
    \caption{(\textbf{a}) Evaluation of the ability of four VLMs to assess the need for patient referral for treatment of wet AMD. (\textbf{b}) Overall referral accuracy was calculated using F1 score for urgent referral with a 95\% CI. Tests of statistical significance were carried out using McNemar's test. The performance of individual ophthalmologists is shown by two white points. (\textbf{c}) Confusion matrices between the senior ophthalmologists assessment (rows) against the image-based clinical decision maker's referral assessment (columns). (\textbf{d}) Image reports written by the non-specialist optician who originally referred the patient, compared with reports of the same patient written by RetinaVLM-Specialist.}
    \label{fig:figure_5}
\end{figure}

As the prevalence of AMD is expected to further increase in the upcoming decades \cite{wong2014global}, ocular screening programs are being introduced around the world. In the United Kingdom, some projects involve opticians and pharmacies that acquire and interpret OCT images. They may refer a patient to a specialist clinic, summarizing their findings and the estimated level of the patient's risk in a letter. In the United Kingdom, treatment guidelines for AMD mandate that patients with signs of neovascularization are referred for immediate treatment within two weeks. However, non-specialists exhibit a tendency to over-diagnose these cases. An internal audit at Southampton Eye Unit found that 74.2\% of the referrals made to the clinic do not have any form of treatable AMD. The processing and assessment of these false positives affects the clinic's ability to care for the remaining patients with treatable forms of AMD.

We evaluated the ability of VLMs to assess the level of referral urgency from OCT image (see \ref{fig:figure_5}a). For each case, the VLMs were provided explicit referral guidelines, and asked to recommend which of three levels of referral urgency was most appropriate for the patient: \textit{no referral} for healthy patients, \textit{to be seen within 18 weeks (routine referral)} for patients that are at risk of progressing to active late wet AMD but do not require treatment yet, and \textit{referral within two weeks} for patients with any signs of neovascularization that should be urgently referred for antiangiogenic treatment. Two junior ophthalmologists independently reviewed images of 95 patients that have previously been referred to the hospital for treatment of wet AMD. For each patient, they independently decided the most appropriate of the three levels of referral urgency, and disagreements were arbitrated by the two senior ophthalmologists. In line with previous audits, they found the false discovery rate for urgent referrals was 69.5\%. 
We then calculated F1 scores for the highest risk patients in need of urgent referral between the VLM's predictions and the ground truth. The full referral protocol and report generation instructions given to the VLMs are provided in Section \ref{sec:method:referral}.

We found that both medical foundation VLMs and Retina-Base perform worse than opticians regarding their ability to refer patients in need of urgent treatment (see Figure \ref{fig:figure_5}b). While Med-Flamingo failed to refer any of the 29 high-risk patients cases, LLaVA-Med and RetinaVLM-Base were ineffective for differentiating high-risk patients from low- to moderate-risk patients (see Figure \ref{fig:figure_5}c). {{} In comparison, ChatGPT-4o was relatively more effective for detecting urgent referrals, but still recommended the referral of 10 patients with low risk, and rarely classified a patient with moderate risk.} RetinaVLM-Specialist was the best peforming VLM, and was able to detect 23 out of the 29 high-risk cases that require immediate treatment. At the same time, RetinaVLM's false discovery rate, defined as the ratio of the number of false positives over the number of predicted positives, of 42.5\% is substantially lower than that of opticians at 69.5\%. Owing to their ability to better differentiate moderate from high-risk cases, the human ophthalmologists had the lowest false discovery rate of 9.1\%, although they simultaneously missed three more cases in urgent need for treatment. { A full table of F1 scores for this task, including a comparison to a standard image-only classification model, are shown in Figure \ref{extended_fig:results_tables}a}.

In practice, referral letters should communicate the reason for referral by citing suspected abnormalities in the OCT image that can inform the ophthalmologist's initial diagnostic plan. As in the conciseness study in Figure \ref{fig:figure_4}, RetinaVLM-Specialist sometimes documents the presence of small biomarkers that cannot be found in the image. More often, RetinaVLM-Specialist wrote an accurate imaging report but did not accurately follow the complex set of referral guidelines provided in the instruction.
This led RetinaVLM-Specialist to incorrectly recommend that 17 of the moderate-risk patients potentially require treatment. However, this occurred less for the 25 low-risk patients, where RetinaVLM-Specialist correctly identified patients with little or no abnormalities, which are often referred to the treatment clinic for a second opinion by non-specialists (samples 1 and 3 in Figure \ref{fig:figure_5}d). 
Crucially, we find that RetinaVLM-Specialist is effective in the detection of intraretinal cysts and fluid that differentiate high-risk from moderate-risk patients (samples 4 and 5). Four more examples of success and failure cases of RetinaVLM-Specialist are shown in Figure \ref{extended_fig:extra_success_failure}b.

\subsection{RetinaVLM accurately detects imaging biomarkers to make recommendations}
\label{sec:results_biomarkers}

\begin{figure}
    \centering
    \includegraphics[width=0.95\linewidth]{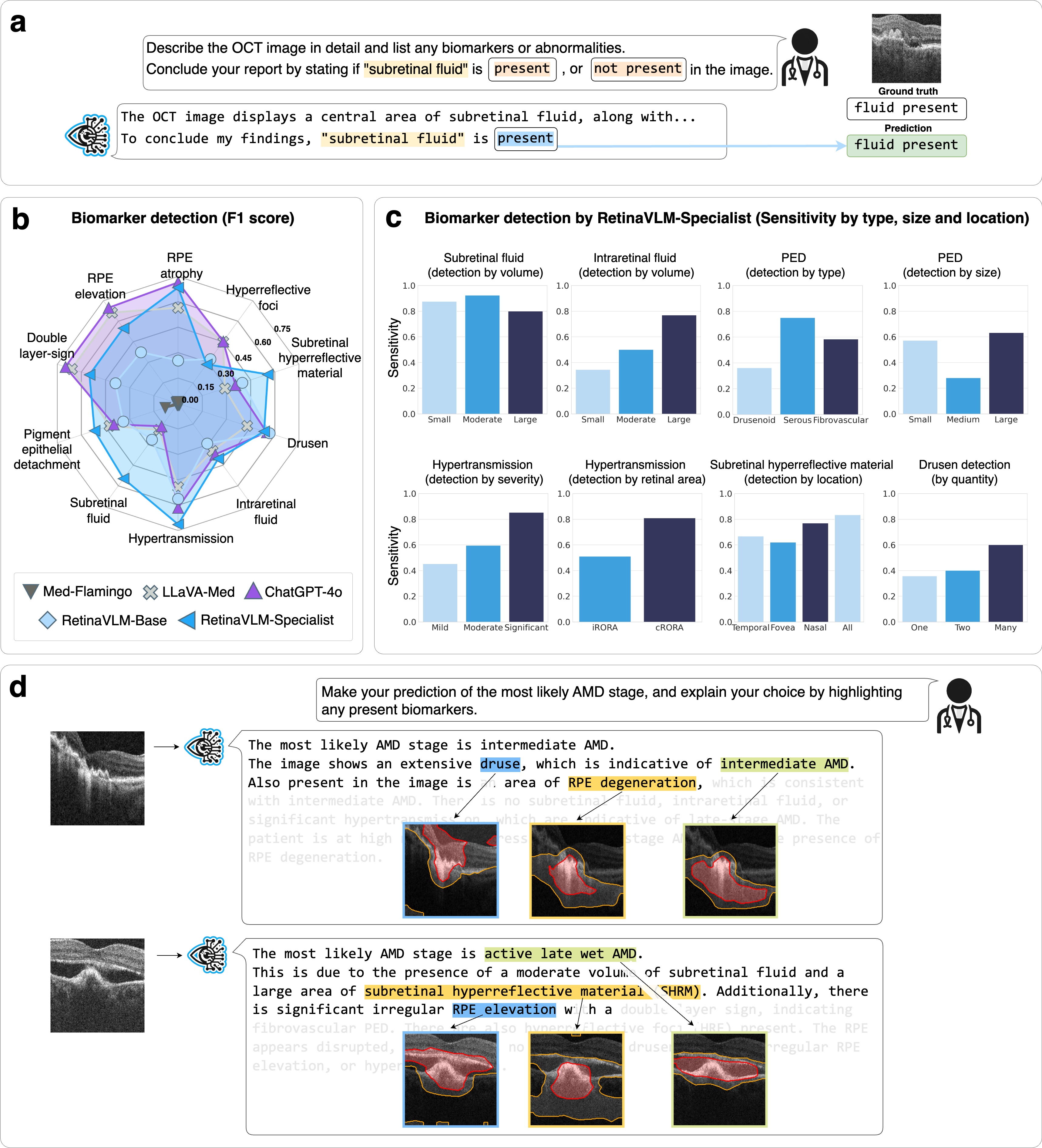}
    \caption{(\textbf{a}) We test four VLMs on their ability to describe the presence of 10 important imaging biomarkers of AMD. (\textbf{b}) Overall detection accuracy was computed using F1 scores. (\textbf{c}) Detection sensitivity for each level of biomarker severity for the most important biomarkers. (\textbf{d}) Regions highlighted by RetinaVLM-Specialist that correspond with different biomarkers and disease stage assessments in its written report. Regions with greater than 25\% and 50\% importance are highlighted by yellow and red contours, respectively. Four additional examples are shown in Figure \ref{extended_fig:gradcam}.}
    \label{fig:figure_6}
\end{figure}

It is important that clinical decision makers can provide evidence for their recommendations. Disease staging reports and written referral recommendations commonly contain descriptions of the most salient biomarkers that were detected in the scan. We tested the ability of four VLMs to correctly identify the presence or absence of 10 different biomarkers related to AMD. To this end, all VLMs were tasked with writing reports for 396 OCT images that conclude by stating the presence or absence of the biomarker in question (see Figure \ref{fig:figure_6}a). The VLMs predictions were compared against the ground truth labels obtained from junior ophthalmologists. The instruction used to generate these biomarker focused reports is provided in Section \ref{sec:method:biomarker_verification}.

We find that RetinaVLM-Specialist outperforms both LLaVA-Med and Med-Flamingo in the detection of seven out of the ten of main biomarkers related to AMD (see Figure \ref{fig:figure_6}b). {{} In overall performance, RetinaVLM-Specialist performed similarly well as ChatGPT-4o. In general,} biomarkers that were more severe, larger, and more numerous were detected with higher accuracy by RetinaVLM-Specialist than less advanced presentations (see Figure \ref{fig:figure_6}c). 
Most of the smaller biomarkers, such as small amounts of intraretinal fluid, drusen and hyperreflective foci, which can be as small as 30 $\mu m$ in size \cite{fragiotta2021significance}, were detected with lower sensitivity.
Overall, clinically important hallmarks of late AMD were detected with a very high sensitivity. Large volumes of subretinal and intraretinal fluid were detected in 80\% and 78\% of cases, respectively, and severe levels of hypertransmission in 84\% of cases. {{} We believe that stronger image encoders capable of detecting smaller features are necessary to improve the performance of RetinaVLM-Specialist on the detection of smaller biomarkers.}

{ 
Finally, we compute saliency maps which highlight regions of the image that were most important to the model in writing specific passages of the report (see Figure \ref{fig:figure_6}d). Qualitatively, we find that RetinaVLM-Specialist attends to relevant regions of the image containing fluid, hypertransmission and retinal pigment epithelium (RPE) irregularities in order to compose passages related to biomarkers and disease stage. Rather than performing biomarker segmentation, for which standard deep segmentation models by \cite{isensee2021nnu} would be a more appropriate tool, these maps provide some explanation as to the choice of words and report passages made by the model. We calculate these using Grad-CAM \cite{selvaraju2017grad} as described in Section \ref{method:saliency_maps}, and provide four additional examples in Figure \ref{extended_fig:gradcam}.}

\section{Discussion}
\label{sec:discussion}

\subsubsection*{Main findings of the study}
{{} In this study we have presented a curriculum-based approach for the specialization of medical vision-language models that directly leverages the knowledge of domain experts. Given a retinal OCT image, the resulting RetinaVLM-Specialist model generates} accurate, detailed textual responses related to disease staging, referral or biomarker identification of AMD. While large foundation deep learning models have been employed for retinal image analysis before \cite{de2018clinically,zhou2023foundation}, our generative VLM is the first model that can flexibly process varied textual queries related to complex ophthalmological decisions and return detailed written responses. Through the use of language as primary communication medium, artificial intelligence systems are able to dynamically perform new tasks and meet the evolving requirements of image-based clinical decision makers. 

In extensive experiments, RetinaVLM-Specialist significantly outperformed {{} ChatGPT-4o and two generative VLMs, Med-Flamingo and LLaVA-Med, designed for medical use. RetinaVLM-Base and ChatGPT-4o were more capable in classifying conversion to late stages of AMD than existing medical foundation models, Flamingo and LLaVA-Med. However, they also lacked the ability to identify subtle yet important differences between disease stages and patient risk. Specifically, we have shown that in disease staging, RetinaVLM-Specialist outperforms all baselines} and is approaching the accuracy of junior ophthalmologists. Similarly, when testing the ability of VLMs to screen for high-risk patients, {{} ChatGPT-4o outperforms non-specialist opticians on aggregate} but significantly underperforms compared to RetinaVLM-Specialist and junior ophthalmologists. In comparison, RetinaVLM-Specialist's reports reduced the number of incorrect urgent referrals by almost four times compared to opticians and had higher recall for urgent referrals than junior ophthalmologists. Finally, RetinaVLM is able to reinforce its decisions by citing observable biomarkers within the written report, and highlighting their corresponding regions within the image.

We postulate that the poor performance of {{} ChatGPT-4o and medical foundation models alike} stems from their lack of detailed knowledge related retinal OCT and AMD. Current VLMs {{} including ChatGPT-4o and LLaVA-Med} are trained on broad, unstructured datasets that are extracted from medical textbooks, scientific publications or social media posts of healthcare professional \cite{li2023llava}. In the United Kingdom and the United States, clinical trainees aspiring to become specialists must undergo up to ten years of post-graduate training to obtain the grade of a board-certified consultant. {{}Classifying AMD requires identifying subtle yet crucial differences between disease stages, particularly in biomarkers such as hypertransmission, drusenoid versus fibrous pigment epithelial detachments, and subretinal hyperreflective material — none of which ChatGPT-4o reported on. While ChatGPT-4o is capable of explaining these differences in language, our analysis indicates it cannot yet make these distinctions in images}. {This reinforces our insight that the paired image-text} training data of current foundation VLMs lacks specialist and experiential knowledge, hindering their effective application to real-world clinical tasks.

A core innovation of our work was the creation of a dedicated training curriculum that specializes VLMs in image-based clinical decision making. Analogously to current medical education, this curriculum deconstructs clinical problems into sets of mandatory capabilities required for their resolution and selectively trains VLMs in these skills. To this end, we obtained a large number of tabular reports by processing of retrospectively collected clinical data using advanced algorithms. Additionally, we tasked ophthalmologists to produce a limited number of highly specific textual reports. In total, our curriculum comprises 41,926 OCT images with 479,710 corresponding visual questions and answers. While still modest in size compared to substantially larger foundation datasets, we believe such curated needs-driven approaches are required to deploy language models specialist healthcare. In a similar vein, leading technology companies in artificial intelligence have also started to look beyond the internet's image and text data to source specialized training data to train LLMs and VLMs in disciplines such as computer programming, journalism, mathematics \cite{sambasivan2021everyone,heikkila2023openai}. {{} Future work may study specializing generalist models such as ChatGPT-4o by finetuning them on specific, high-quality curricula such as the one introduced in this study.}


\subsubsection*{Limitations and future research directions}
\label{sec:discussion_limitations}

{
Naturally, the quality of our curriculum depends on the underlying imaging reports, and in particular those used to create curriculum part 2. The majority of the reports used to create RetinaVLM-Specialist were written by a junior ophthalmologist with three years experience. The remaining reports were written by an ophthalmologist with ten years of experience, and their reports were more comprehensive. 
While reports written by both were found to be of high quality, future work could evaluate the benefits of a third training curriculum derived by senior ophthalmologists.

By updating instruction sets, VLMs have the potential to be more adaptable than image-only deep learning approaches to differences in clinical practice between countries. However, it is important to note that our curriculum was derived using UK-specific clinical definitions and workflows. As a result, the errors made by RetinaVLM were found to reflect differences in the biomarker and staging definitions used by the UK-based and Austria-based ophthalmologists involved in the testing of the model. In particular, the model was found to be conservative in classifying fibrovascular features that upgrade intermediate AMD to inactive late wet AMD. Thus, to match or surpass the performance of the average junior ophthalmologist, we also aim to establish a consistent cohort of ophthalmologists for training and testing the model.}

Similarly, we observed a discrepancy in image interpretation between junior and senior ophthalmologists. Junior ophthalmologists did not recommend patients for referral if it was likely that the retinal fluid observed was caused by traction rather than neovascularization, as it is not treatable with antiangiogenic drugs. Conversely, the senior ophthalmologist preferred that these patients be still referred for immediate assessment to rule out neovascularization. RetinaVLM was explicitly instructed to refer patients with any sign of fluid of any cause (see Section \ref{sec:method:referral}), and correctly referred more patients as a result.

Another technical limitation of our approach was the sensitivity of the LLM generating the question-answer pairs to the specifics of its instructions. Extensive trial and error were required to arrive at several instructions, listed in Section \ref{supplemental:qa_prompts}, that resulted in diverse sets of high quality question-answer pairs. We discern that all aspects of dataset creation - deciding on the required capabilities, collecting specialized annotations and converting these to question-answer pairs - should be formalized to systematically compare different approaches and ultimately scale dataset curation in the future.

RetinaVLM also inherits some of the fundamental limitations of language models. LLMs are prone to confidently present false or fabricated information, termed hallucinations, which has been identified as problematic in medical contexts \cite{singhal2023large,hager2024evaluating}. { This phenomenon has also been observed in VLMs, where the generated text does not relate to any object or feature that is observable in the image \cite{li2023evaluating,liu2024survey}.} Similarly, we observed that RetinaVLM occasionally hallucinates the presence of retinal fluid and consequently diagnoses more advanced AMD stages than necessary. RetinaVLM's output was also sensitive to the wording of questions and instructions. While this had little impact on our qualitative analysis, extensive trial and error was necessary to ensure that RetinaVLM responded with one of the provided options in the quantitative analyses.

Beyond the formalization and extension of the curriculum, there are other potential technical improvements to RetinaVLM. Currently, RetinaVLM processes a single two-dimensional OCT image from one type of OCT scanner. In ophthalmological practice, decisions are made based on three-dimensional images from multiple time points, although many recent studies on the use of foundation models in ophthalmology also analyze two-dimensional images \cite{zhou2023foundation}. We mitigated the impact of this discrepancy on our study by tasking ophthalmologists to select the most relevant two-dimensional slice of the imaged volume before proceeding with the referral decision. In the future, a more sophisticated vision encoder, which is able to handle three-dimensional data from diverse OCT imaging devices, could be integrated with RetinaVLM. {{} Our results also indicate that stronger image encoders capable of detecting smaller features are necessary to improve the performance of RetinaVLM-Specialist on the detection of smaller biomarkers.} Similarly, one may opt to incorporate multimodal information, such as health questionnaires, clinical tests or the patient's medical history, into the decision making process \cite{baltruvsaitis2018multimodal}. The fundamental model architecture and training would remain similar, but the level of reasoning required for differential diagnosis across multiple scans would potentially increase.

Similarly, we exclusively trained RetinaVLM for the management of a single retinal disease, AMD, ignoring other retinal pathologies such as diabetic retinopathy or glaucoma, or imaging modalities, such as color fundus photography \cite{flaxman2017global,abramoff2010retinal}.  While this enabled us to explore the potential to encode advanced clinical levels of specialism into VLMs at depth, it would severely limit the applicability of the current version of our models for general ocular screening. It also led the model to classify all cases of retinal fluid as wet AMD, where a few cases showed evidence of diabetic retinopathy and other retinal conditions.

{ In order to ready VLMs for deployment for ocular screening, future work would need to extend RetinaVLM's curriculum with domain experts from an expanded range of ophthalmological conditions. This requires costly and time-consuming curation of specialized training datasets by medical experts. However, we believe that this is a necessary investment as routine clinical skills and patient management protocols are rarely documented in existing datasets used to train AI models.}

\subsubsection*{Conclusion}

Foundation vision-language models (VLMs) have the potential to revolutionize healthcare by automatically interpreting medical images and communicating their findings in detailed written reports. Trained on large datasets containing millions of medical images and textual annotations, foundation language models have stirred considerable interest for their expert-level performance on medical licensing exams and case studies. However, in this work we have shown that foundation medical VLMs {{} such as ChatGPT-4o} substantially underperform compared with human experts on routine clinical tasks.

We hypothesize that the training data of foundation medical VLMs currently lacks specialist clinical knowledge and experience. To address this, we developed a curriculum-based approach that integrates the expertise of domain specialists into the training of medical VLMs. The resulting model, RetinaVLM, can produce detailed imaging reports that make accurate recommendations for the clinical management of AMD. It approaches and often matches the performance of junior ophthalmologists in disease staging, and outperforms non-specialist opticians in patient referral.

These results indicate that merely increasing the scale of training datasets is insufficient for the development of VLMs with real-world clinical utility. Instead, medical VLMs require high-quality data directly related to the challenges faced by clinicians in their daily practice. We believe our proposed curriculum-based approach provides a blueprint for specializing VLMs that generate true value in healthcare.

\clearpage
\section{Extended figures}

\begin{figure}
    \centering
    \includegraphics[width=0.95\linewidth]{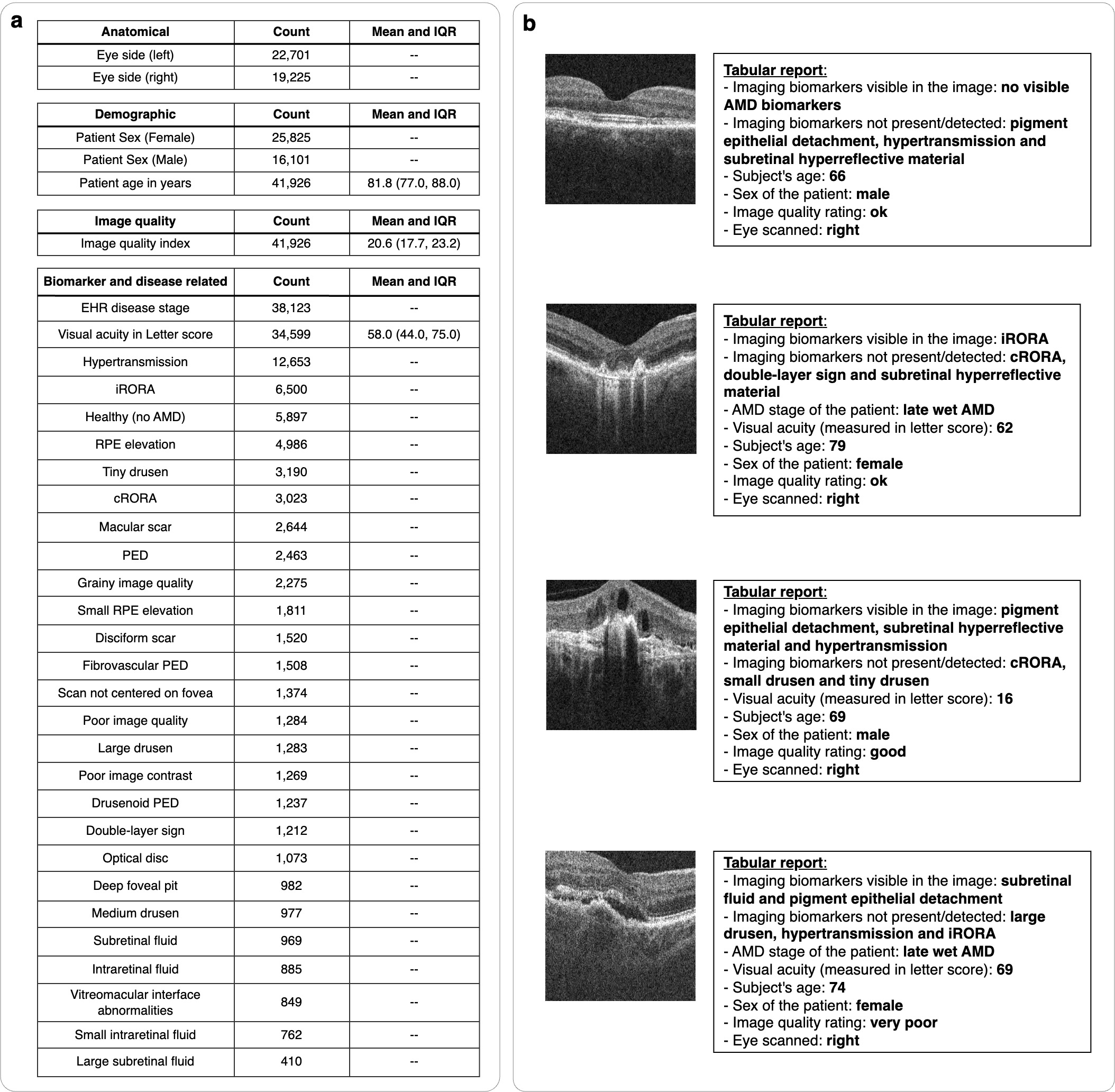} 
    \caption{ {{} (\textbf{a}) The count of each tabular label used to create the tabular reports for the training portion of curriculum part 1. Labels are divided into those relating to anatomy, demography, quality, and biomarkers and disease. By default, variables are not mutually exclusive (an image may show drusen and subretinal fluid), with the exception of those introduced with a bracketed notation, such as Eye side (left) and Eye side (right). We report mean and inter-quartile bounds for continuous-valued variables, and leave this blank for categorical variables} (\textbf{b}) Four randomly selected images and their corresponding tabular reports in text form.}
    \label{extended_fig:tabular_reports}
\end{figure}

\begin{figure}
    \centering
    \includegraphics[width=0.95\linewidth]{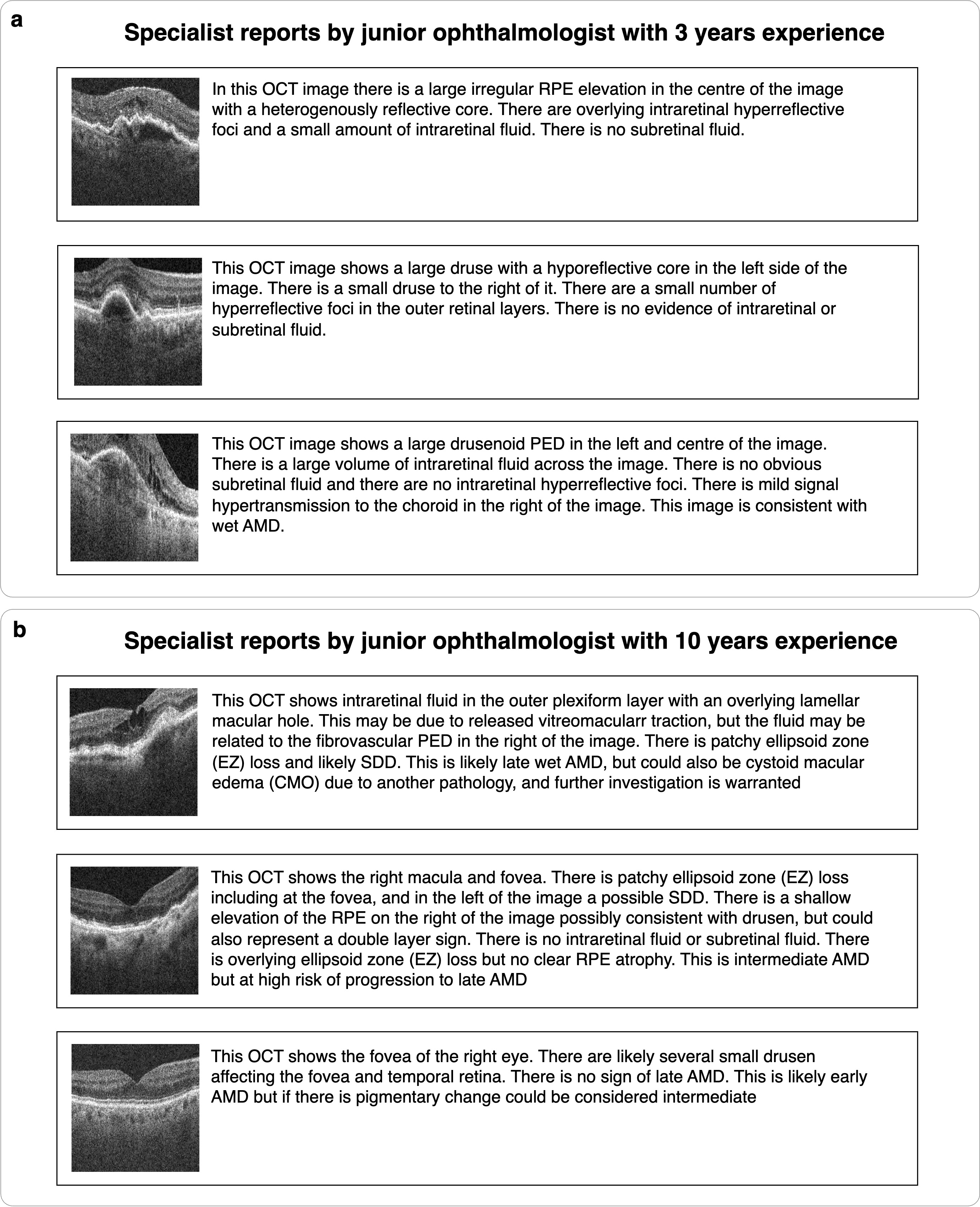}
    \caption{(\textbf{a}) Three example reports randomly selected from the 244 written by the first junior ophthalmologist with 3 years experience in ophthalmology. (\textbf{b}) Three example reports randomly selected from the 86 written by the more experienced junior ophthalmologist with 10 years experience.}
    \label{extended_fig:specialist_reports}
\end{figure}

\begin{figure}
    \centering
    \includegraphics[width=0.92\linewidth]{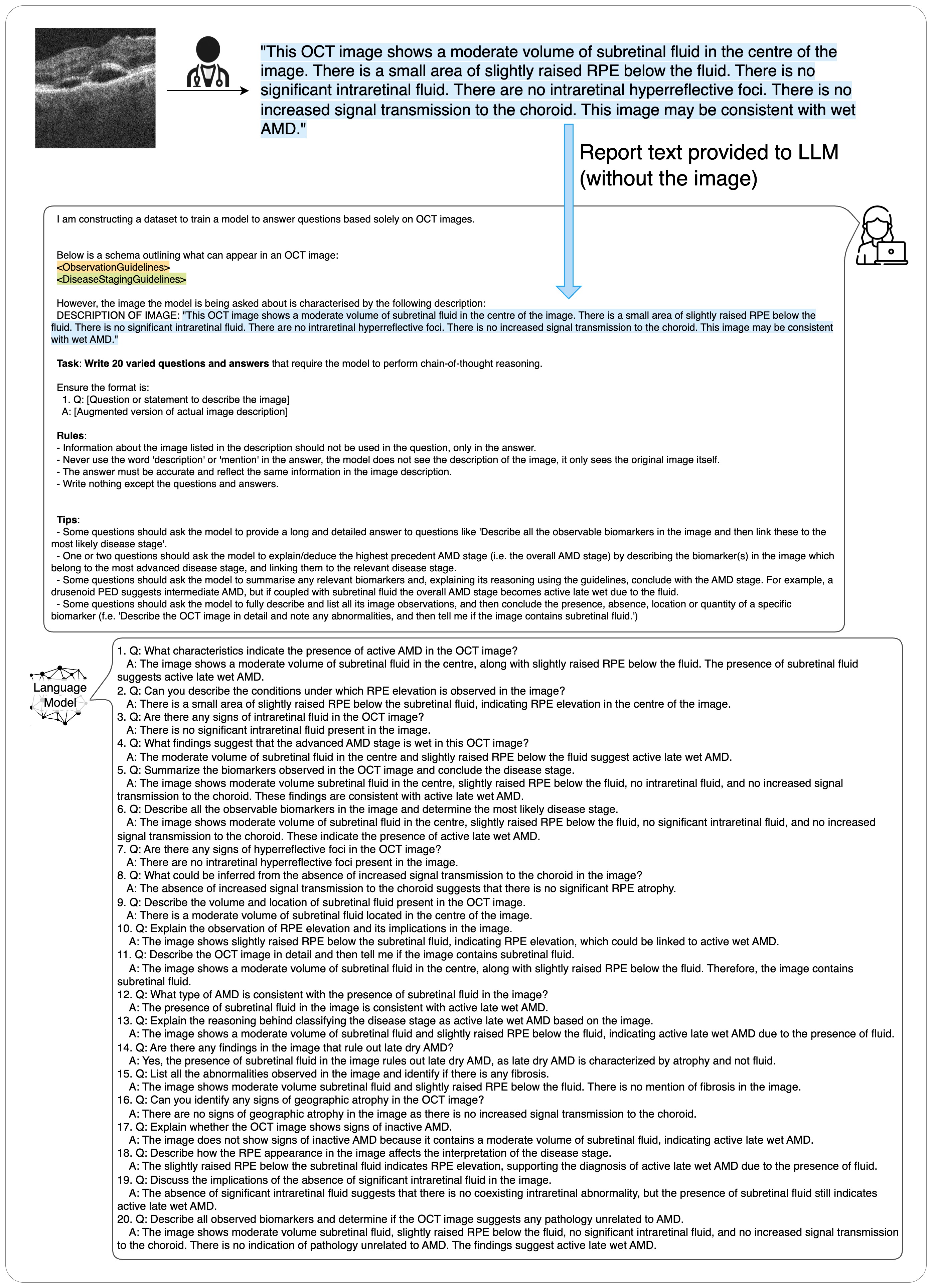}
    \caption{Example interaction with the LLM used to generate the question-answer pairs. Given the report text, the LLM is instructed to generate a set number of question-answer pairs in a specific format, according to pre-determined rules, and the guidelines derived by ophthalmologists. References to guidelines are replaced with the corresponding guideline text, documented in Figure \ref{extended_fig:curricula}.}
    \label{extended_fig:qa_generation}
\end{figure}

\begin{figure}
    \centering
    \includegraphics[width=0.92\linewidth]{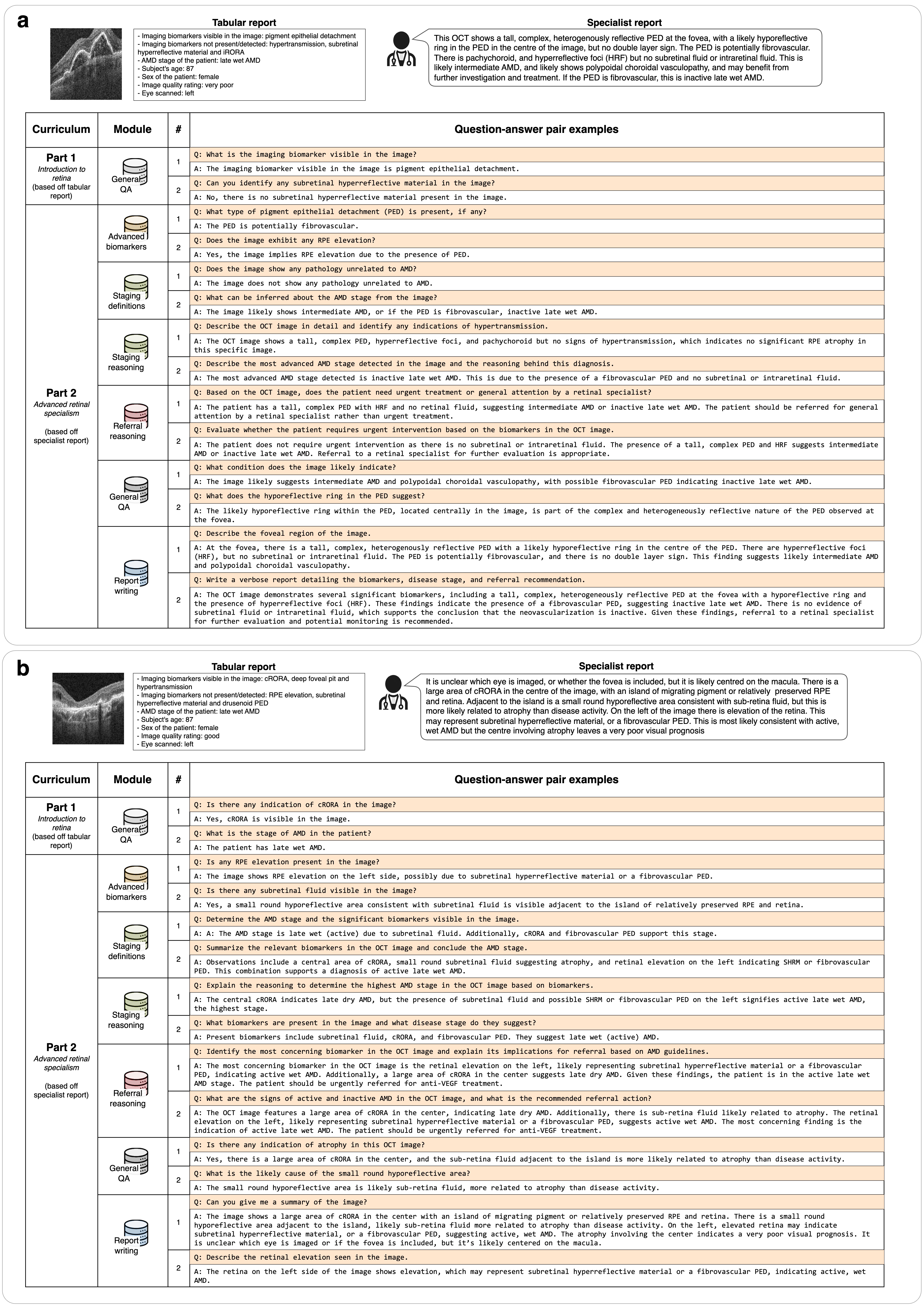}
    \caption{(\textbf{a}) Example question-answer pairs generated for curriculum part 1 (based on the tabular report), and for the six modules that constitute part 2 (based on the specialist report). (\textbf{b}) Example question-answer pairs for a second example image, derived from different tabular and specialist reports.}
    \label{extended_fig:qa_examples}
\end{figure}

\begin{figure}
    \centering
    \includegraphics[width=0.95\linewidth]{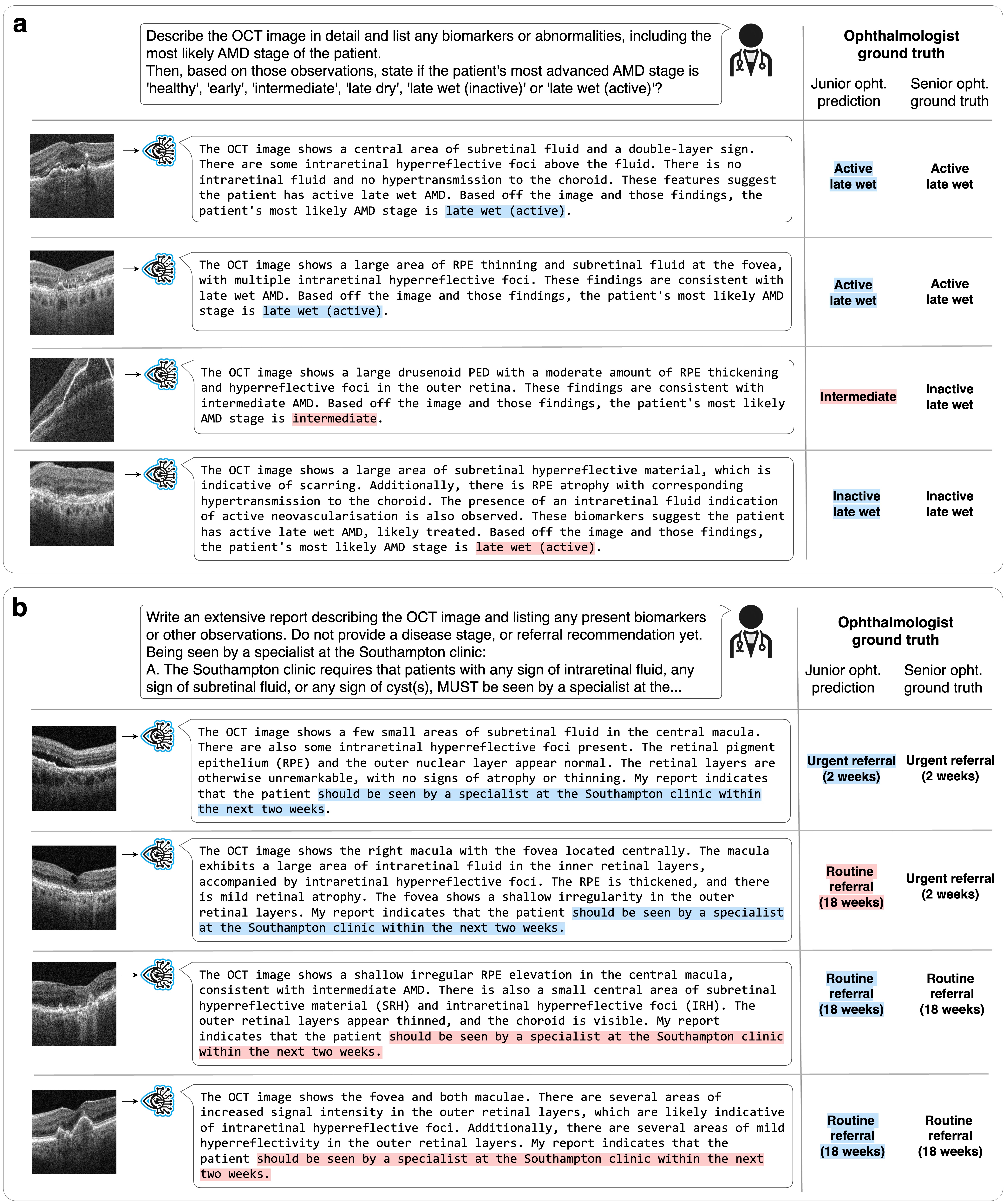}
    \caption{(\textbf{a}) Four example reports by RetinaVLM-Specialist that result in two correct and two incorrect disease stage conclusions. Images showing inactive late wet AMD sometimes had fewer abnormalities and were misclassified as intermediate AMD (third sample). Another mode of failure was the hallucination of retinal fluid, which unnecessarily upgraded the disease classification to active late wet AMD (fourth sample). (\textbf{b}) Four example reports that result in two correct and two incorrect patient referral recommendations. In each failure case RetinaVLM-Specialist correctly identifies the lack of fluid, but fails to follow the referral protocol and makes the wrong referral recommendation, which was the most common mode of failure for this task.}
    \label{extended_fig:extra_success_failure}
\end{figure}

\begin{figure}
    \centering
    \includegraphics[width=0.95\linewidth]{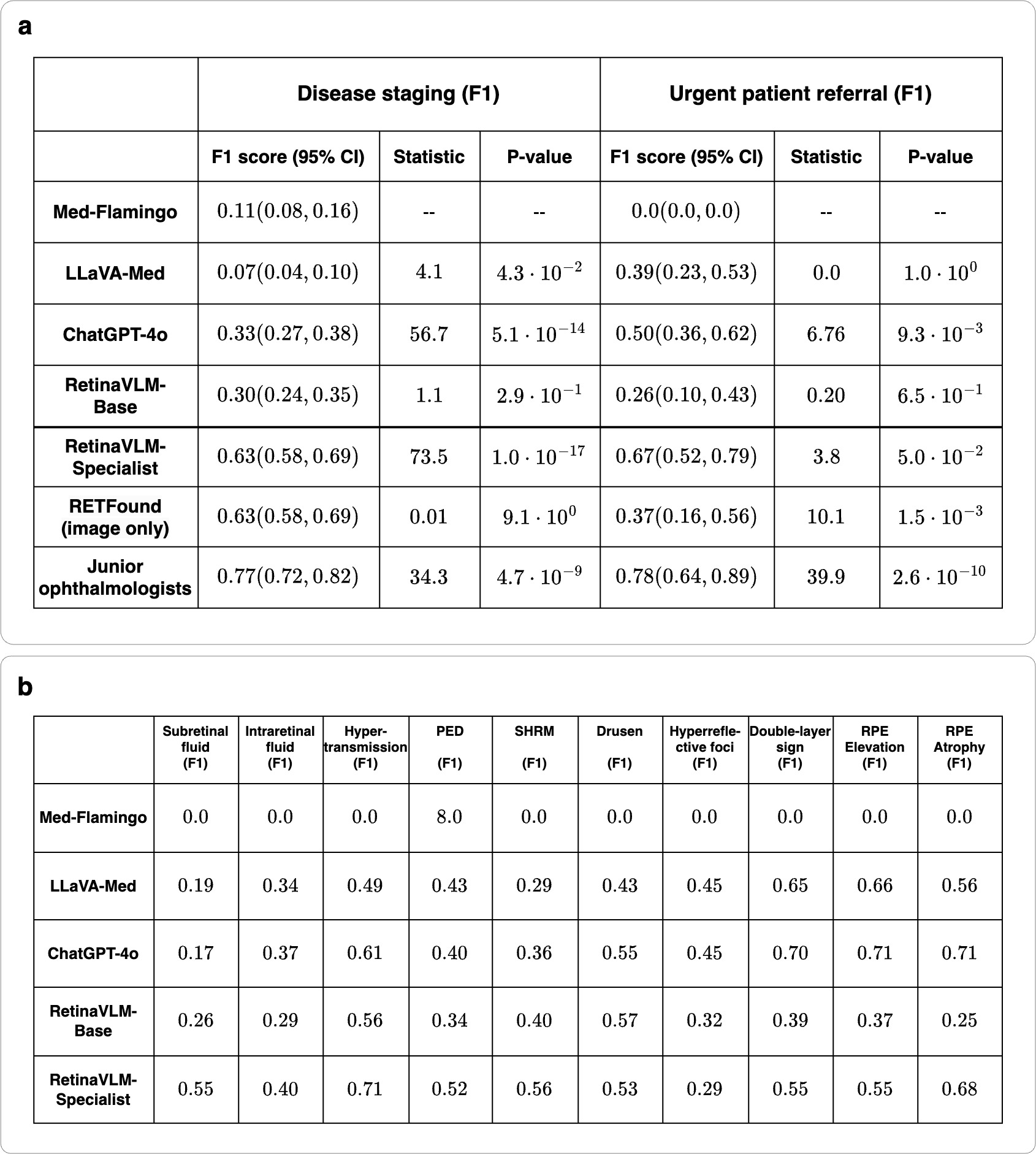}
    \caption{{ (\textbf{a}) Full results tables for the disease staging and patient referral tasks. Significance tests are made between successive models in order, with the exception that the RETFound image-only baseline and the junior ophthalmologists were both compared against RetinaVLM-Specialist. Details of the RETFound image-only baseline are documented in Section \ref{method:image_only_baseline}. (\textbf{b}) Full results table for the detection of ten different biomarkers related to AMD.}}
    \label{extended_fig:results_tables}
\end{figure}

\begin{figure}
    \centering
    \includegraphics[width=0.95\linewidth]{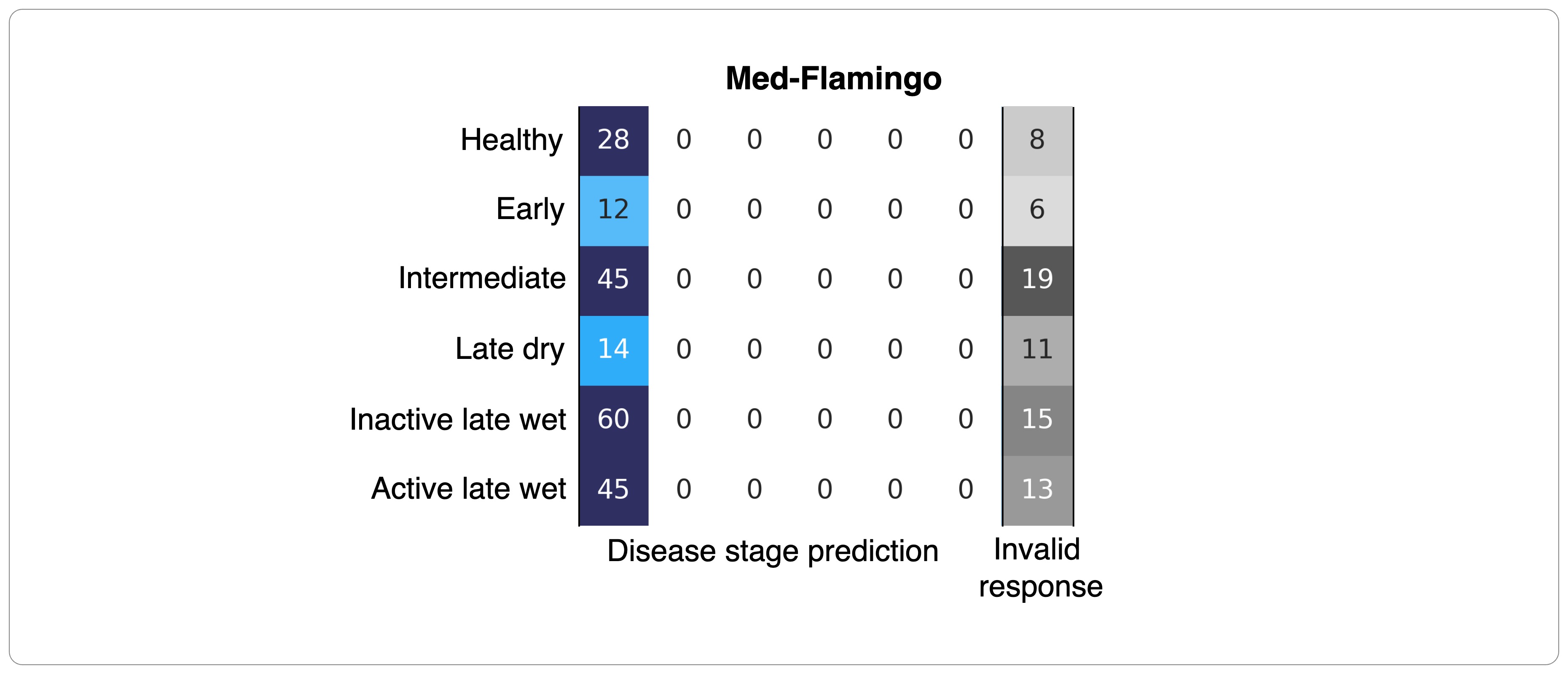}
    \caption{Confusion matrix for Med-Flamingo, included here for space reasons. Med-Flamingo always predicted that the patient was healthy irrespective of the true disease state of the image or returned an invalid report.}
    \label{extended_fig:med-flamingo_confusion}
\end{figure}

\begin{figure}
    \centering
    \includegraphics[width=0.95\linewidth]{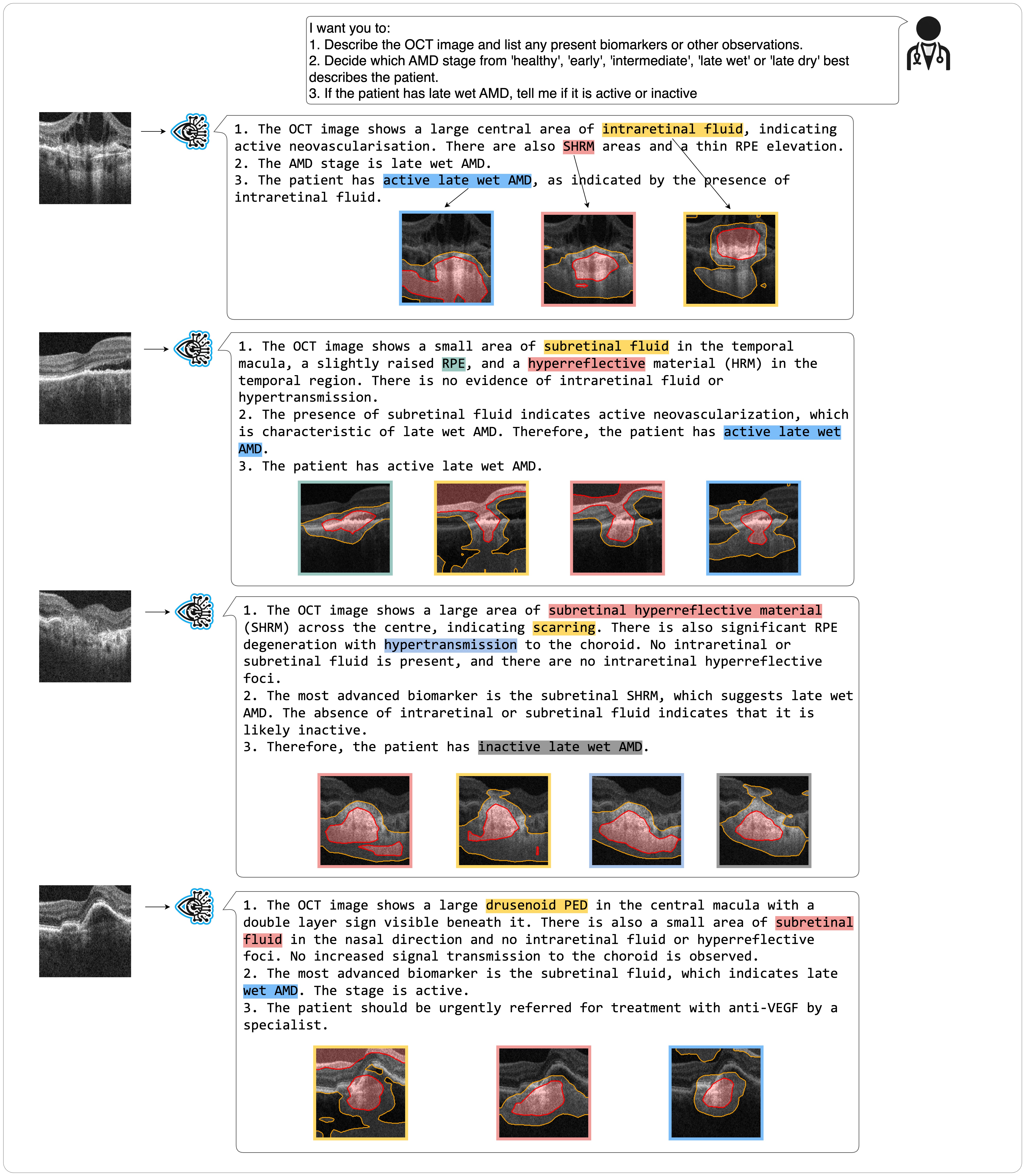}
    \caption{Additional examples of saliency maps computed with RetinaVLM-Specialist. Each saliency map is generated for specific phrases of the report, where color indicates correspondence. Saliency maps related to fluid often contain the fluid boundary. Saliency maps sometimes include unrelated regions, such as for subretinal fluid in the second sample, and the pigment epithelial detachment (PED) in the fourth sample. Moreover, in the last sample the model misidentifies the presence of subretinal fluid and the saliency map is centered on hypertransmission. Overall, the saliency maps assist in highlighting more severe biomarkers, and in interpreting the working of the model.}
    \label{extended_fig:gradcam}
\end{figure}

\begin{figure}
    \centering
    \includegraphics[width=0.95\linewidth]{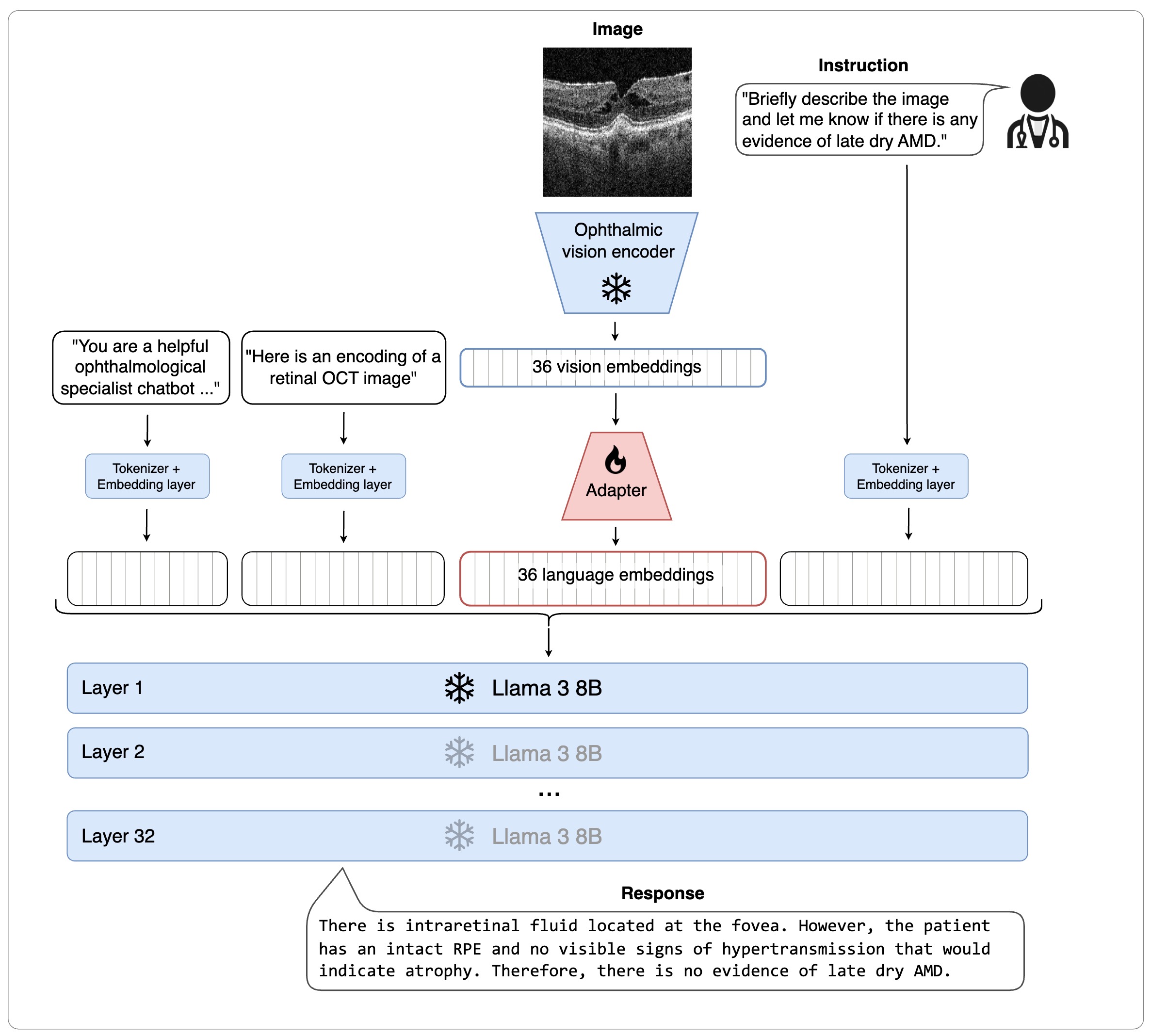}
    \caption{RetinaVLM consists of an ophthalmic vision encoder, a LLM and an image-to-language adapter. The vision encoder and LLM are not updated during training. Textual instructions are projected to the embedding space of the LLM using the pre-existing tokenizer and embedding layer, while the vision embeddings are projected via the learnable adapter. During training, the adapter is optimized such the the LLM gives the corresponding response to the given instruction.}
    \label{extended_fig:architecture}
\end{figure}

\begin{figure}
    \centering
    \includegraphics[width=0.95\linewidth]{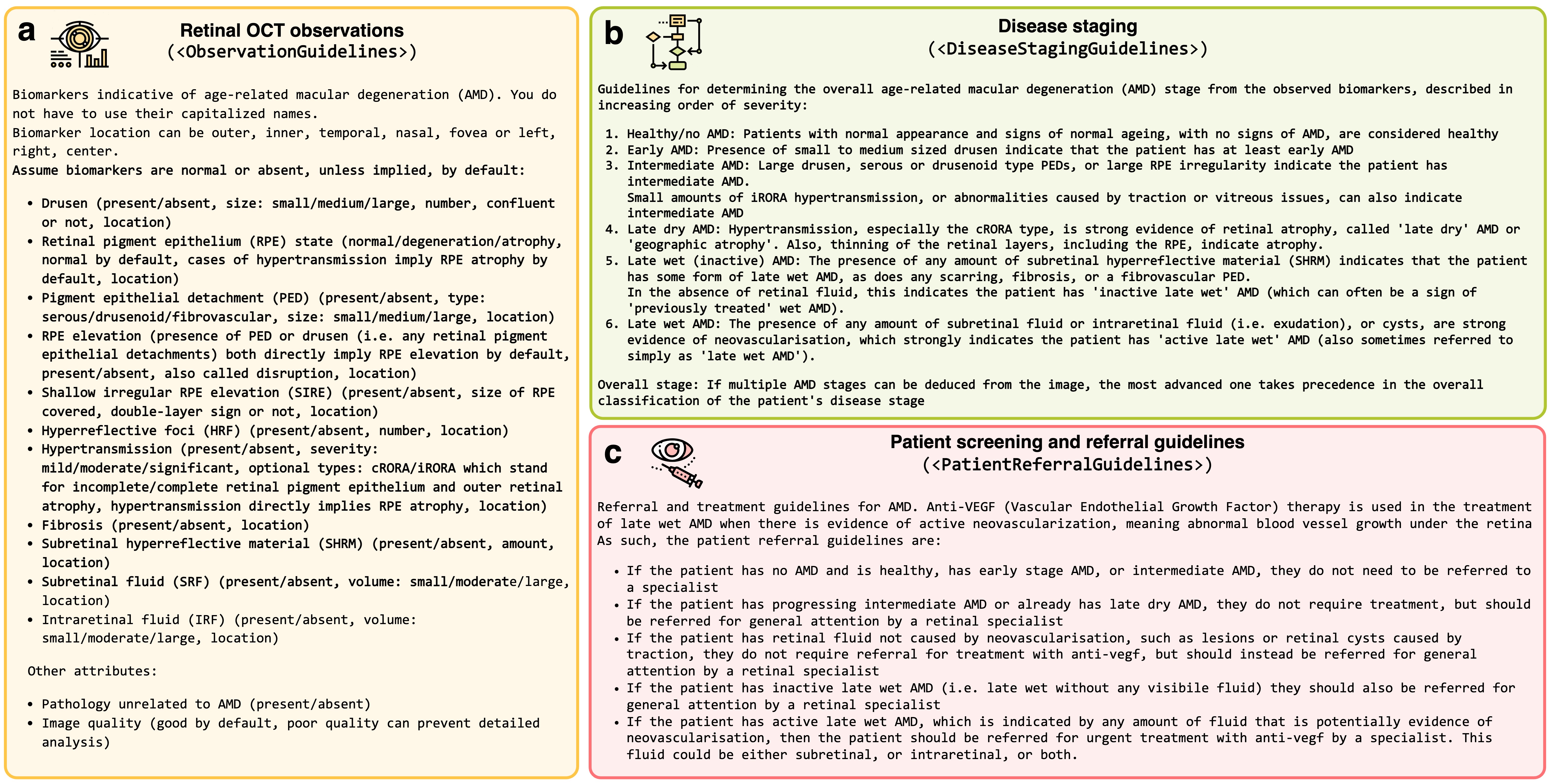}
    \caption{Three guidelines defined by junior and senior ophthalmologists for the assessment and referral of AMD from retinal OCT. All can be downloaded from our open-source repository (see Section \ref{sec:method:code_availability}).  These guidelines are used to guide the LLM generating question-answer pairs that train VLMs in the mandatory capabilities of image-based clinical decision makers for AMD. (\textbf{a}) Biomarkers observable in retinal OCT that are relevant to AMD. (\textbf{b}) Linking observable biomarkers to deduce progressively severe disease stages. (\textbf{c}) Protocol for patient referral for treatment with antiangiogenic drugs and routine referral, based on the observations and disease stage.}
    \label{extended_fig:curricula}
\end{figure}

\clearpage
\section{Methods}
\subsection{Retinal image dataset curation}

We use two retinal OCT datasets in this study. The first, described in Section \ref{sec:method:oct_dataset}, contains a cohort of patients with AMD collected retrospectively at the Southampton Eye Unit. The second dataset, described in Section \ref{sec:method:patient_referral_dataset}, contains scans of the initial visits of patients referred, primarily by opticians, to the Southampton Eye Unit. { The curation and use of this data is summarized in a flowchart diagram in Figure \ref{fig:data_curation}}.

All data was collected in the scope of the PINNACLE study (ClinicalTrials.gov NCT04269304), which received approval from the East Midlands–Leicester Central Research Ethics Committee in the United Kingdom (ref. 19/EM/0163) and the institutional review boards of all participating institutions. 
It complies with the principles of Good Clinical Practice and the Declaration of Helsinki. All images were captured using Topcon 3D OCT scanners (Topcon Corporation, Tokyo, Japan). Both datasets contain images of size $416\times512$ with a pixel size of 3.5$\times$11.7 $\mu m^2$.

\subsubsection{Retrospective dataset for training RetinaVLM and testing disease staging, report writing and biomarker analysis}
\label{sec:method:oct_dataset}

The retrospective dataset contains 44,633 OCT images from 6,152 eyes belonging to 3,468 patients, collected over eight years, between 2012 and 2020, at the Southampton Eye Unit and curated by the PINNACLE consortium. For each OCT scan we use the mediolateral 2D slice centered at the fovea.

{ We then designated 41,926 of the 44,633 OCT images from 5,547 eyes of 3,057 patients for training purposes.} Additionally, we reserved 2,311 images from 326 eyes of 187 patients for validation, and 396 images from 279 eyes of 224 patients for testing. We ensured that images from each patient do not appear in more than one of the training, validation or test sets. The training set was used to create both curriculum parts 1 and 2, detailed in Sections \ref{sec:method:tabular} and \ref{sec:method:specialist}. The test set was used to evaluate the resulting model in Sections \ref{sec:results_staging} and \ref{sec:results_biomarkers}. 

{{} In Section \ref{sec:results_staging}, we introduced our multi-rater process for curating high-quality labels to evaluate the accuracy of image report conclusions. We now expand on this by detailing the breakdown of disease stage definitions used in this study. AMD remains a relatively poorly understood disease, with multiple grading systems that vary in stage classification and definitions  \cite{bird1995international,klein2014harmonizing,ferris2005simplified,ferris2013clinical,sadda2018consensus}. Its classification remains an active area of research \cite{sadda2018consensus}. Classifying a retinal OCT image into a single disease stage is challenging, as overlapping features can indicate multiple AMD stages, often requiring ophthalmologists to make intuitive assessments beyond strict criteria. In this study, the ophthalmologists reported classifying healthy and early AMD based on the presence or absence of small drusen, and intermediate AMD by medium, or intermediate, to large drusen \cite{klein2014harmonizing}. Late dry AMD was identified by atrophy of the retinal pigment epithelium, evidenced by hypertransmission. Presence of any subretinal hyperreflective material or other scarring advanced the classification to inactive wet AMD. Finally, presence of any subretinal or intraretinal fluid upgraded the image-based classification to active late wet AMD. These classifications yielded a testing set of 276 images that labeled 36 as healthy, 18 as early, 64 as intermediate, 25 as late dry, 75 as inactive late wet and 58 as active late wet AMD.

Each image was labeled with the presence or absence of 10 biomarkers, and an additional 21 labels that record their size, number, and other applicable attributes. Due to the substantially increased number of variables compared to disease staging, each image was labeled only once by one of the six junior ophthalmologists. This process yielded a testing set of 396 images that labeled 107 as evidencing drusen, 70 with pigment epithelial detachments, 150 with pigment epithelial elevation, 27 with double-layer sign, 122 with hypertransmission, 155 with atrophic/degenerated pigment epithelium layers, 60 with subretinal hyperreflective material, 81 with hyperreflective foci, 25 with subretinal fluid, and 62 with intraretinal fluid.
}

\subsubsection{External testing dataset of patients referred to Southampton AMD treatment clinic}
\label{sec:method:patient_referral_dataset}
We also collected an external dataset of 95 patients that were referred primarily by opticians for urgent care at the Southampton Eye Unit between 02/2023 and 12/2023. None had yet received treatment for AMD, and mostly had no AMD, intermediate AMD or small features related to active wet AMD. This represents a distribution shift from the retrospective cohort, where many patients had already received treatment for AMD and were in the inactive late wet stage of AMD. As such, it enabled us to estimate the robustness of both variants of RetinaVLM to shifts in patient population. This dataset was not used for model training and was reserved for testing VLMs on patient referral, detailed in Section \ref{sec:results_referral}.

{{} For each patient we sourced scans of both their left and right eye that were acquired on their first visit to the clinic. We also collected the originally issued letter of referral, as depicted in Figure \ref{fig:figure_5}d. Then, two junior ophthalmologists independently reviewed the 3D OCT volumes of both eyes to label the patient's risk and decide from three levels of increasing referral referral urgency. To help standardize their labels, the ophthalmologists referred to a set of agreed patient referral guidelines. For full documentation see \texttt{<PatientReferralGuidelines>} in Figure \ref{extended_fig:curricula})c.

After labeling the level of referral urgency, the ophthalmologists then selected the image slice that most supported their assessment of the patient's risk. In healthy patients where both volumes contained no pathological signs in any of the image slices, they were instructed to select the mediolateral fovea-centered 2D slice from one of the two volumes. Finally, any inter-rater disagreements were then arbitrated by a panel involving the two senior ophthalmologists. Of the 95 images in the external cohort, 25 did not evidence need of referral, 41 indicated the patient was at moderate risk and needed general attention by a specialist, while only 29 indicated the patient needed urgent referral.}

 {
\subsection{RetinaVLM vision-language model architecture}
\label{method:vlm_model}
RetinaVLM follows the architectural design of MiniGPT4 \cite{zhu2023minigpt}. It consists of two main components: an ophthalmological vision encoder and a generative LLM (see Figure \ref{extended_fig:architecture}). For the ophthalmological vision encoder, we adopt a Resnet50 convolutional neural network with over 23 million parameters which was pre-trained with self-supervised contrastive learning on the 41,926 OCT images from the train set of the retrospective cohort. Specifically, it was trained with Bootstrap Your Own Latent (BYOL) \cite{grill2020bootstrap} using the same implementation details as the standard contrastive approaches in \cite{holland2024metadata}, which consistently performed on par with RETFound \cite{zhou2023foundation} on data from the Southampton Eye Unit.} This vision encoder projects each $192 \times 192$ input image to a set of spatially arranged 6$\times$6 visual embeddings, which are extracted from the last layer before global average pooling. Each embedding has a dimension of $h_{img}=2048$. They also have a receptive field of size 336, so each embedding contains global knowledge of the image that is contextualized at its local position. 

For the LLM, we employ the 8 billion parameter instruction-tuned Llama3 model by Meta \cite{MetaLlama3,llama3modelcard} as the generative LLM, which was was the most performant openly available model at the time of our study. LLama3 uses an embedding dimension of $h_{lang}=4096$.

{ The ophthalmological vision encoder provides visual information regarding the OCT image to the LLM via an adapter. The adapter is a linear layer of size $h_{img} \times h_{lang}$ that processes visual information for use by the LLM. Specifically, it does so by independently mapping each of the visual embeddings, applying a linear transformation via matrix multiplication, into language embeddings used by the LLM. The design and application of the adapter follows the design used in MiniGPT4 \cite{zhu2023minigpt}, and results in an adapter with over 8 million parameters.}

\subsection{Foundation vision-language models}
We used the two most widely adopted foundation vision-language models for medical applications at the time of this study \cite{moor2023med,li2023llava}. They were both trained on large biomedical datasets sourced from the Internet, and have been applied in chest x-ray \cite{chen2024chexagent}. The first, Med-Flamingo \cite{moor2023med}, which was built on Flamingo \cite{alayrac2022flamingo} and finetuned on image and text data from medical textbooks and the PubMed Central Open Access (PMC-OA) dataset \cite{lin2023pmc}. The second, LLaVA-Med \cite{li2023llava}, developed by Microsoft, is a VLM built on LLaVA \cite{liu2024visual} and finetuned to follow textual instructions regarding a broad range of biomedical images contained in PubMed Central 15M (PMC-15M) \cite{zhang2023large}. { The training sets of both models contain retinal OCT images and associated text. As they were trained as generalist models on various imaging modalities, they were both purportedly capable of interpreting retinal OCT images. When provided a retinal OCT image we found both could correctly identify its modality and begin generating diagnoses, without being provided any prior information.} {{} More recently, other generative vision-language models for ophthalmology have been introduced but were either not designed for OCT imaging nor finetuned with instruction-tuning \cite{deng2024ophglm,zhang2024generalist}}. {{} Our third baseline constitutes OpenAI’s GPT-4o model (endpoint ‘gpt-4o-2024-05-13’). Unlike the two aforementioned medical VLMs, GPT-4o is a generalist model.}

For Med-Flamingo, we then provide instructions using the following template provided in their code, replacing \texttt{\{question\}} with the instruction text:

\begin{quote}
\begin{Verbatim}[breaklines, baselinestretch=1.0, fontsize=\fontsize{11}{12}\selectfont, breaksymbolleft=, breaksymbolright=]
You are a helpful medical assistant. You are being provided with images, a question about the image and an answer. Follow the examples and answer the last question. <image>Question: {question} Answer:
\end{Verbatim}
\end{quote}

{ Similarly, for LLaVA-Med we use the following conversation template provided by the developers:}

\begin{quote}
\begin{Verbatim}[breaklines, baselinestretch=1.0, fontsize=\fontsize{11}{12}\selectfont, breaksymbolleft=, breaksymbolright=]
A chat between a curious human and an artificial intelligence assistant. The assistant gives helpful, detailed, and polite answers to the human's questions.###Human: {question}###Assistant:
\end{Verbatim}
\end{quote}

{{} Similarly, for the ChatGPT-4o API we provide the following system prompt:

\begin{quote}
\begin{Verbatim}[breaklines, baselinestretch=1.0, fontsize=\fontsize{11}{12}\selectfont, breaksymbolleft=, breaksymbolright=]
You are an intelligent and helpful assistant. You follow all the instructions in the user prompt, and answer all the questions they ask for.
\end{Verbatim}
\end{quote}
}

\subsection{Report curation and question-answer pair generation}

\subsubsection{Curriculum part 1: Introduction to retina}
\label{sec:method:tabular}
To create the tabular reports for the first part of the curriculum we used a cluster-based approach to efficiently label the 41,926 training images with biomarker annotations \cite{holland2024deep}. {{} This training cohort includes 25,825 female and 16,101 male patients with an average age of 81.8 years, with lower (Q1) and upper (Q3) quartiles of of 77.0 and 88.0 years.}

Contrastive learning is used to extract self-supervised features from the dataset. The dataset is then partitioned into 40 clusters of images that share common features. Labels are then assigned to these clusters by senior ophthalmologists. To this end, 20 images from each cluster were reviewed by senior ophthalmologists. If the majority of the images exhibited common features, such as 'large drusen' or 'subretinal fluid', these labels were assigned to the entire cluster. These labels were used in in combination with the patient's age, sex and their functional visual acuity score (measured on a LogMAR chart and converted to Letter score) to create the tabular reports. { We also included the quality index of the image using an intensity-histogram-based approach \cite{stein2006new}. We found that labeling the dataset's quality index quartiles as `very poor', `ok,' `good,' and `excellent' effectively captured the characteristics of each quartile.} Additionally, the reports list three biomarkers that are stated as not being present. These are drawn from a distribution of all biomarkers, weighted by their prevalence in the dataset, that were not among the cluster labels for that image. {{} Counts of the prevalence of each tabular variable among the training images are shown in Figure \ref{extended_fig:tabular_reports}a and b, and a sample of four tabular reports they result in are shown in Figure \ref{extended_fig:tabular_reports}c.}

{{} \noindent \paragraph{Creating visual question-answer pairs from tabular reports:} To generate question-answer pairs from the large volume of tabular reports we used an LLM available for download and local use. We chose WizardLLM-70B, though any freely-available LLM with instruction-following capabilities may be used for this purpose. This model was chosen as an open-weights model with strong performance in instruction following and general knowledge at the time of dataset creation.} We then used the instruction template detailed in Section \ref{supplemental:curriculum_part1}. This resulted in a total of 408,505 question-answer pairs, or an average of 9.75 question-answer pairs per report depending on its complexity. Examples of the question-answers pairs generated by this approach are shown in the `curriculum part 1' section of Figure \ref{extended_fig:qa_examples}a and Figure \ref{extended_fig:qa_examples}b.

\subsubsection{Curriculum part 2: Advanced retinal specialism}
\label{sec:method:specialist}

The second part of the curriculum involves further turning on a subset of 330 images from the training curriculum in part 1. {{} This training cohort includes 205 female and 125 male patients with an average age of 80.8 years, with lower (Q1) and upper (Q3) quartiles of of 75.0 and 87.0 years.

Unlike curriculum part 1, images used in curriculum part 2 were annotated with high quality reports manually curated by retinal specialists.} Specifically, two junior ophthalmologists were tasked with describing the main pathological biomarkers and diagnoses related to AMD, while also noting any other observation regarding the retinal anatomy. This yielded high-quality textual reports that go beyond the short notes that ophthalmologists typically write in clinical routine. The first junior ophthalmologist, with three years of experience specializing in ophthalmology, wrote the majority of 244 reports (see Figure \ref{extended_fig:specialist_reports}a). While these were highly accurate, they were less comprehensive in their analysis than the remaining 86 reports written by the junior ophthalmologist with 10 years of experience (see Figure \ref{extended_fig:specialist_reports}b).

Simultaneously, the same two junior ophthalmologists used the same methodology to produce 28 reports on images from the test set. In our analysis, these were found to be of high quality by the two senior ophthalmologists (see Figure \ref{fig:figure_4}). As they were representative of the 330 reports collected on the training set, the senior ophthalmologists concluded that these results provide sufficient quality assurance for the reports used to generate curriculum part 2.

{{} \noindent \paragraph{Creating visual question-answer pairs from specialist reports:} After the imaging reports were curated, we used an external LLM with 10 different instructions to generate up to 230 questions per image.} The exact instructions used are documented in Section \ref{supplemental:qa_prompts}. We take two preliminary steps before providing the instruction to the LLM. We firstly replace the \texttt{<ReportText>} identifier with the raw text of the image report. Additionally, many of the QA generation instructions make references to the guidelines that describe the mandatory capabilities of image-based clinical decision makers with regard to disease staging and patient referral for patients with AMD. The second step involves replacing any reference to the \texttt{<ObservationGuidelines>}, \texttt{<DiseaseStagingGuidelines>} or \texttt{<PatientReferralGuidelines>} by the text of the corresponding guidelines (documented in Figure \ref{extended_fig:curricula}). These guidelines were verified by senior ophthalmologists, and were instrumental for the generation of questions with improved diversity and coverage, and also for creating questions about biomarkers that were absent in the image (and typically not mentioned in the report).

The smaller number of reports in the advanced curriculum permitted the use of the more performant proprietary models for generating question-answer pairs. We used the `gpt-4o' API endpoint from OpenAI. An example interaction with 'gpt-4o' for generating these question-answer pairs is shown in Figure \ref{extended_fig:qa_generation}. A sample of question-answers yielded by this approach are shown in the `part 2' section of both Figure \ref{extended_fig:qa_examples}a and Figure \ref{extended_fig:qa_examples}b.

{
\subsection{RetinaVLM model inference and training}
\label{method:vlm_training}

\subsubsection{Inference and loss computation}

\label{method:vlm_training:training}
RetinaVLM processes retinal OCT images and textual instructions. To begin building the input provided to the model, we first set the system prompt of the constituent LLM to:
\begin{Verbatim}[breaklines, baselinestretch=1.0, fontsize=\fontsize{11}{12}\selectfont, breaksymbolleft=, breaksymbolright=]
    You are a helpful ophthalmological specialist chatbot capable of interpreting retinal OCT images.
\end{Verbatim}

We then begin the instruction that will be provided to the LLM with the following line:

\begin{Verbatim}[breaklines, baselinestretch=1.0, fontsize=\fontsize{11}{12}\selectfont, breaksymbolleft=, breaksymbolright=]
    Here is an encoding of a retinal OCT image <Img><ImageHere></Img>\n
\end{Verbatim}

For each image in the batch, we next add the text of the specific question to the instruction. For example:

\begin{Verbatim}[breaklines, baselinestretch=1.0, fontsize=\fontsize{11}{12}\selectfont, breaksymbolleft=, breaksymbolright=]
    Describe the OCT image in detail. Does it show evidence of retinal fluid?
\end{Verbatim}

To complete the textual input to the model, we populate the LLM's conversation template with this instruction and include the question's answer exclusively during training. This forms the full textual input, which we project to the embedding space of the LLM using its pre-existing tokenizer and pretrained embedding layer.

After embedding the input text, we process the retinal OCT image in question. To this end we downsample the image by a factor of 2 from $416\times512$ to $208\times256$ pixels. We then crop the image centrally in testing, and augment each image using the finetuning protocol outlined in \cite{holland2024metadata} in training. This results in a processed OCT image of size $192 \times 192$.

Next, we use the ophthalmic vision encoder $E$ to extract the $6\times6$ vision feature embeddings from the image. After flattening these, we apply the adapter to each embedding separately. This projects the visual embeddings to the input embedding space of the LLM. To create the final set of embeddings that are provided to the LLM, we replace the textual embeddings corresponding to the \texttt{<ImageHere>} phrase with the 36 adapted vision embeddings from the actual retinal OCT image.

Finally, the resulting sequence of language embeddings and adapted visual embeddings are passed together through the frozen LLM. This yields a list of predicted token logits with the same length as the input sequence. We then compute the causal language modeling loss between the predicted answer logits and the ground truth answer tokens. We then optimize the adapter to minimize this loss.

Crucially, when training RetinaVLM we keep both the vision encoder and LLM \textit{frozen}. That is, they are not updated during the entire training process.
This is key to preserving RetinaVLM's inheritance of the pretrained LLM's language and reasoning capabilities. These enable RetinaVLM to handle versatile questions and instructions beyond those encountered in the curriculum during training. Thus, during training, we only update the adapter that feeds visual information regarding the OCT image to the LLM.
}

{
\subsubsection{Training on curriculum parts 1 and 2}

RetinaVLM is trained sequentially on curriculum part 1 and part 2. After randomly initializing the adapter we first train the model on the 408,545 question-answer pairs regarding the 41,926 images in curriculum part 1 (introduction to retina) to obtain \textit{RetinaVLM-Base}. The then continue to finetune the model on the 71,165 questions and answers regarding the 330 images in curriculum part 2 (advanced retinal specialism), resulting in the final \textit{RetinaVLM-Specialist} model. For each of the curriculum parts we train RetinaVLM for 100,000 steps using a batch size of 12. In each step we randomly select one question-answer pair per image from the current curriculum. To update the network we use the AdamW optimizer with a learning rate of $10^{-4}$, and $\beta_1 = 0.9$ and $\beta_2 = 0.999$.

Training on the two curriculum parts separately is important for assessing the accumulative benefits of progressively specialist training in two ways. Firstly, this enables us to measure the benefits of training on free-text medical reports over tabular data. Secondly, during development we found that reserving the highest quality data for the final training stage, curriculum part 2, improved the performance of the resulting model. This strategy is standard practice in the development of LLM-based models \cite{wei2021finetuned}.
}

\subsection{Using VLMs for inference and generating text}
After VLMs have been trained on image and text datasets, they can be used to generate responses to new questions and images. To use the baseline foundation VLMs for testing we take the central crop in each $416\times512$ OCT image, resulting in an image of size $384\times384$. {{} After repeating this image along the color dimension, ChatGPT-4o accepts the resulting $3\times384\times384$ RGB image, while the Med-Flamingo and LLaVA-Med baselines require we downsample this to $3\times224\times224$. To provide the image to the RetinaVLM variants, we first downsample the $416\times512$ image by a factor of two to $208\times256$, before taking a central crop resulting in an image of size $192\times192$.}

We then employ the same method with all VLMs for generating responses to instructions. Provided the image and textual instruction, we build the output sequence of tokens by repeatedly appending the token to the output assigned the highest probability by the VLM. This is equivalent to using a temperature parameter set to 0, and is the standard approach for generating the most accurate and least creative output from LLM-based models. { All other generation parameters, such as repetition penalty, were not used}. This process is repeated until a stop token is generated, signaling the end of the VLM's response. The model's tokenizer is then used to convert the numeric output tokens to the final free text output.

\subsection{Experimental setup}
Our entire evaluation was conducted in \textit{zero-shot}, that is, after training on curriculum part 1 and part 2 RetinaVLM requires no further finetuning in order to perform tasks related to disease staging, patient referral and biomarker analysis. Instead, for each test we designed a specific instruction that was provided to all VLMs to generate the application-specific reports that were used in our analyses. { These instructions were derived through experimentation with all VLMs on the validation set. While they are not designed for any one VLM in particular, they do contain information related to the task at hand. To inform their design, we assessed the quality of each prompt by assessing the quality of outputs it produced in all VLMs on a subset of 30 images from the validation set. This led to two observations. Firstly, that all VLMs produced improved responses when prompted to first detail its ‘chain-of-thought’ \cite{wei2022chain}. This led us to request that the model first describe the OCT image in detail before making any recommendations. Secondly, that simpler, more direct prompts resulted in all VLMs making fewer errors in following our subsequent, test-specific instructions.}

\subsubsection{Tests of disease staging}
\label{sec:method:disease_staging}
The following instruction was given to all VLMs to and obtain the reports of the 276 test images that were analyzed in Section \ref{sec:results_staging}. The instruction requests VLMs to begin by describing the image, and then deduce the most advanced disease stage:

\begin{Verbatim}[breaklines, baselinestretch=1.0, fontsize=\fontsize{11}{12}\selectfont, breaksymbolleft=, breaksymbolright=]
    Describe the OCT image in detail and list any biomarkers or abnormalities, including the most likely AMD stage of the patient.
    
    Then, based on those observations, state if the patient's most advanced AMD stage is 'healthy', 'early', 'intermediate', 'late dry', 'late wet (inactive)' or 'late wet (active)'?
\end{Verbatim}

After the VLM generated its report (using a maximum of 500 tokens), we appended the following incomplete sentence to its output:

\begin{Verbatim}[breaklines, baselinestretch=1.0, fontsize=\fontsize{11}{12}\selectfont, breaksymbolleft=, breaksymbolright=]
Based off the image and those findings, the patient's most advanced AMD stage is
\end{Verbatim}

which the VLM them completes using up to 300 tokens. From these tokens, we extracted the final disease staging prediction by searching for the first instance of any of the listed disease stages. This post-processing step is only necessary for quantitative tests of accuracy, as it enables the reliable extraction of the disease stage from the free text report. In cases where the VLM discusses multiple disease stages, such as in `\texttt{more advanced than early AMD, and is intermediate AMD as there is no evidence of late wet AMD}', the disease stage was manually extracted. In cases where no disease stage was provided or could be extracted manually, this counted as an `Invalid response'.

\subsubsection{Evaluations of correctness, completeness and conciseness by senior ophthalmologists}
\label{sec:method:report_ccc}
For the direct evaluation by the senior ophthalmologists, we randomly selected 28 of the test images from the retrospective dataset, and tasked the two junior ophthalmologists with annotating the images. We provided the following instruction to RetinaVLM-Specialist and LLaVA-Med to generate their image reports:

\begin{Verbatim}[breaklines, baselinestretch=1.0, fontsize=\fontsize{11}{12}\selectfont, breaksymbolleft=, breaksymbolright=]
    Write an extensive report describing the OCT image, noting any biomarkers or abnormalities related to AMD, and their qualities. Also comment on which biomarkers are absent. 
    Finally, based on the image and these findings, your report should estimate the AMD disease stage of the patient. 
    You should not include any patient referral recommendations in your report, but you can comment if they need treatment with anti-vegf.
\end{Verbatim}

{{} We found that ChatGPT-4o tended to write excessively long reports. To improve the performance of ChatGPT-4o in direct evaluations by senior ophthalmologists, we requested a `\texttt{brief}' rather than an `\texttt{extensive}' report, and appended the following text to the original instruction: `\texttt{Your report should be no longer than 120 words long}'. This helped make reports more concise with no observable loss in accuracy, though they still exceeded the length of reports written by junior ophthalmologists and RetinaVLM-Specialist.}

We then assigned 14 of the images and the corresponding 42 reports by {{} ChatGPT-4o}, RetinaVLM-Specialist and the junior ophthalmologists, to each of the two senior ophthalmologists, who evaluated them independently according to the correctness, completeness and conciseness. These criteria, also documented in Figure \ref{fig:figure_4}, were:

\let\labelitemi\labelitemii
\begin{itemize}
    \item Correctness - The report is accurate in its main observations and conclusions regarding the image
    \item Completeness - The report contains all relevant observations and conclusions that can be inferred from the image
    \item Conciseness - The report does not make observations and conclusions that are not supported by, or not seen in, the image
\end{itemize}

\subsubsection{Tests of patient referral}
\label{sec:method:referral}
The following instruction was given to all VLMs to generate reports that focus on patient referral recommendations, which are analyzed in Section \ref{sec:results_staging}. This instruction was run for the 95 referral images, introduced in Section \ref{sec:method:patient_referral_dataset}. In order to accurately convey the specific requirements of the wet AMD treatment clinic, we provided the comprehensive referral protocol used by the senior ophthalmologists in the instruction:

\begin{Verbatim}[breaklines, baselinestretch=1.0, fontsize=\fontsize{11}{12}\selectfont, breaksymbolleft=, breaksymbolright=]
    Write an extensive report describing the OCT image and listing any
    present biomarkers or other observations. Do not provide a disease
    stage, or referral recommendation yet.

    Being seen by a specialist at the Southampton clinic:
    A. The Southampton clinic requires that patients with any sign of 
       intraretinal fluid, any sign of subretinal fluid, or any sign of cyst(s), MUST be seen by a specialist at the Southampton clinic within the next two weeks.
    B. The Southampton clinic requires that patients who do not have any sign of 
       intraretinal fluid, any sign of subretinal fluid, or any sign of cyst(s), but do have some biomarkers of early or intermediate AMD, should be seen by a specialist at the Southampton clinic for routine referral.
    C. The Southampton clinic requires that patients who do not have any sign of 
       intraretinal fluid, any sign of subretinal fluid, or any sign of cyst(s), but do have medium to large drusen, drusenoid PED, hypertransmission or atrophy, should be seen by a specialist at the Southampton clinic for routine referral.
    D. The Southampton clinic does not need to see patients who have no biomarkers 
       and healthy retinas at all.

    Southampton specialist visit: Next, tell me if your initial report of the OCT 
    image indicates that the patient should be seen by a specialist at the 
    Southampton clinic "within the next two weeks", to be seen "within 18 weeks 
    (routine referral)", or "not be seen" at all?
\end{Verbatim}

As before, after the VLM generated its report (using a maximum of 500 tokens), we added the following incomplete sentence to its output:

\begin{Verbatim}[breaklines, baselinestretch=1.0, fontsize=\fontsize{11}{12}\selectfont, breaksymbolleft=, breaksymbolright=]
    My report indicates that the patient
\end{Verbatim}

which is then completed by the VLM for up to a maximum of 300 tokens. We then searched these tokens for the first instance of one of the three levels of referral urgency, `\texttt{within the next two weeks}', `\texttt{within 18 weeks (routine referral)}' or `\texttt{not be seen}', in the VLM's output report. { In cases where different wording is used, but with identical meaning, the VLM's prediction is extracted manually. In the remainder of cases, it is listed as an `Invalid response'.}

\subsubsection{Tests of biomarker analysis}
\label{sec:method:biomarker_verification}

The following instruction was given to all VLMs to generate reports that conclude the presence of absence of the 10 different biomarkers, evaluated on the 396 test images in Section \ref{sec:results_biomarkers}:

\begin{Verbatim}[breaklines, baselinestretch=1.0, fontsize=\fontsize{11}{12}\selectfont, breaksymbolleft=, breaksymbolright=]
    Describe the OCT image in detail and list all biomarkers or abnormalities. 
    Detail if there are any signs indicating that {biomarker} might be present, 
    even if there is only a small amount.

    Finally, conclude your findings by telling me if {biomarker} {article} "not 
    present", or if potentially any amount of {biomarker} {article} "present" in 
    the OCT image.
\end{Verbatim}

For each of the 10 biomarkers, the phrase \texttt{\{biomarker\}} was replaced by the actual biomarker name (such as `subretinal fluid'), and the \texttt{\{article\}} replaced by \texttt{`is'} for singular biomarkers or \texttt{`are'} for plural biomarkers (such as drusen). After the VLM generated its report (using a maximum of 500 tokens), we appended the following incomplete sentence to its output:

\begin{Verbatim}[breaklines, baselinestretch=1.0, fontsize=\fontsize{11}{12}\selectfont, breaksymbolleft=, breaksymbolright=]
    To conclude these findings, in the OCT image {biomarker} {article}
\end{Verbatim}

which the VLM then completes using up to 300 tokens. We then searched for the first instance of \texttt{not present} or \texttt{present} to extract the model's prediction of the absence or presence of the biomarker, respectively. { In cases where different wording is used, but with identical meaning, such as stating the biomarker was `\texttt{detected}' rather than `\texttt{present}', the VLM's prediction is extracted manually. In the remainder of cases, it is listed as an `Invalid response'.}

\subsubsection{Computing language-based image saliency maps}
\label{method:saliency_maps}
We provide methodogolical details for the computation of the language-based saliency maps discussed in Section \ref{sec:results_biomarkers}, and shown in Figure \ref{fig:figure_6}d and Figure \ref{extended_fig:gradcam}. With saliency maps we aim to identify which regions of the image were most relevant to certain passages, such as \texttt{large subretinal fluid}, of RetinaVLM-Specialist's responses. The most direct way to generate these visualizations to use attention maps, but we found Llama3's pretrained attention maps did not result in any meaningful saliency maps. To address this, we used Grad-CAM \cite{selvaraju2017grad}, a technique for highlighting the most relevant image regions to the prediction of an image classifier. By defining the predicted class as the sum over the tokens in the output passage, which formulates the LLM as an image classifier, we were able to generate the saliency maps shown in Figure \ref{fig:figure_6}. A code implementation can be found at the repository referenced in Section \ref{sec:method:code_availability}.

{
\subsection{Comparison to an image-only classification-based deep learning baseline}
\label{method:image_only_baseline}

VLMs are an emerging technology that hold great potential to automate language-based reporting and decisions regarding medical images. To provide a comparison between VLMs and more established approaches to classification we use RETFound, which was pretrained on over 700,000 2D OCT images and exhibits strong performance when finetuned on as few as 100 images \cite{zhou2023foundation,holland2024metadata}. RETFound makes predictions from the image alone and cannot interpret nor respond to written language. As an image encoder, RETFound instead outputs a single classification per image.

We compare the VLMs against RETFound in both the disease staging and patient referral tasks.
To this end, we formulated disease staging as a six-way classification task using the same stages as the experiment shown in Figure \ref{fig:figure_3}. Similarly, we formulated the patient referral task as a three-way classification task using the same referral levels as the experiment shown in Figure \ref{fig:figure_4}. In both cases, we manually extract training labels from the imaging reports of the same 330 training images used in curriculum part 2 to train RetinaVLM-Specialist (see Section \ref{sec:advanced_specialism}). 

After formulating these tasks, we used the standard linear evaluation approach to evaluate the performance of the RETFound imaging encoder. This involves applying global average pooling to RETFound's output tokens to yield a feature vector of size $1024$. We then train a linear layer that maps this feature vector to the output classes of the given classification problem. This was done using an AdamW optimizer with learning rate of $3 \cdot 10^{-4}$ for $5000$ training steps. In both cases, performance converged and little to no overfit was observed on the validation set. We then selected the step with the best performance on the validation set for evaluation. Finally, this classifier is tested on the same arbitrated test labels as all VLMs in Figures \ref{fig:figure_3} and \ref{fig:figure_4}.

The results of both experiments are shown in Figure \ref{extended_fig:results_tables}a. For the disease staging task we find that RETFound (0.63 F1) significantly outperforms the best performing foundation medical VLMs (0.11 F1) and RetinaVLM-Base (0.30 F1). Notably, it performs as well as RetinaVLM-Specialist (0.63 F1). However, on the patient referral task RETFound (0.37 F1) performs on par with the best foundation VLM (0.39 F1) and significantly worse than RetinaVLM-Specialist (0.67 F1).

The equal performance of RetinaVLM-Specialist and the image-only RETFound baseline in disease staging implies that both have reached an upper bound on performance for this task on the retrospective dataset. 
However, we observed relatively poor performance from the image-only baseline on the patient referral task, which we attribute to domain shift. 
Specifically, the retrospective dataset lacks images with intact retinal pigment epithelium layers that feature small fluid pockets, which represent many of the urgent referral cases in the referral dataset. This image-only model's ability to generalize well to these cases.
While accurate patient referral is achievable using an image-only encoder like RETFound, addressing this domain shift would necessitate the collection of a new training dataset that is more representative of the referral cohort.
In contrast, RetinaVLM-Specialist is better able to mitigate this domain shift by utilizing task-specific textual instructions that provide details about the target domain — namely, the patient referral protocol outlined in Section \ref{sec:method:referral}.

}

\subsection{Measurements of performance and statistical analysis}
{ To calculate the performance of each VLM and retinal specialist in all multiple-choice question answering, we used the micro F1 score.} This aggregates the total number of false positives (FP), false negatives (FN), true negatives (TN) and true positives (TP) over all classes before computing the F1 score using equation \ref{eq:f1}

\begin{equation}
F_1 = \frac{2 \cdot TP}{2 \cdot TP + FP + FN}
\label{eq:f1}
\end{equation}

{ In cases where the VLM returned an `Invalid response' by failing to pick one of the listed options this was counted as a false negative for the ground truth class.} After calculating the F1 score, we determined the 95\% confidence interval through bootstrapping $N=1000$ times with replacement. 

Tests of significance (aggregated in Figure \ref{extended_fig:results_tables}) were calculated using a two-sided McNemar's test \cite{mcnemar1947note}. This test assesses the difference in the number of correctly versus incorrectly predicted samples, focusing on cases where the models agree or disagree on the labels. A significant p-value from the McNemar test allows us to reject the null hypothesis that both models have identical classification performance. We then used the following notation to indicate levels of statistical significance: *** for $p \leq 0.001$, ** for $p \leq 0.01$, and * for $p \leq 0.05$ and `ns' (not significant) for $p > 0.05$.

\subsection{Computing hardware and software}
We use Python 3.12.2 to conduct all model question-answer generation, VLM training, and VLM evaluation. To generate the question-answer pairs for curriculum part 1 we used 3 40GB NVIDIA A40 GPUs. For both training RetinaVLM and for evaluating all VLMs we use a single 80GB NVIDIA A100 GPU and PyTorch \cite{paszke2017automatic} version 2.1.2. Training RetinaVLM on takes 1 day on curriculum part 1, and another day on curriculum part 2. Llama3 was downloaded via Huggingface with model ID `meta-llama/Meta-Llama-3-8B-Instruct'. The baseline VLM Med-Flamingo's code and model weights were installed following the instructions at \url{https://github.com/fastscience-ai/MedFlamingo}, and LLaVA-Med's from \url{https://github.com/microsoft/LLaVA-Med}. Confusion matrices and results calculations were computed with scikit-learn version 1.4.1 and numpy version 1.26.4. Figures and tables were created in draw.io v24.4.0 using plots generated by matplotlib version 3.8.4 and seaborn version 0.13.1. Grad-CAM was computed using grad-cam version 1.5.0. McNemar's tests of significance were calculated using statsmodels version 0.14.1.

\section{Data availability}
Both imaging datasets are currently being curated and maintained by the Vienna Reading Center on behalf of the PINNACLE consortium. The data will be made available to once the PINNACLE study concludes in 2026 \cite{sutton2023developing}.

\section{Code availability}
\label{sec:method:code_availability}
{{} The code and guidelines used to create the question-answer pairs, train, and evaluate the models can be found at \href{https://github.com/RobbieHolland/SpecialistVLMs}{https://github.com/RobbieHolland/SpecialistVLMs}. The code may be used to develop the models can be repurposed for other medical specialties. Moreover, we make all model weights for RetinaVLM-Base and RetinaVLM-Specialist openly available at \href{https://huggingface.co/RobbieHolland/RetinaVLM}{https://huggingface.co/RobbieHolland/RetinaVLM}. These are only intended for research purposes related to retinal OCT images.}

{{}
\section{Acknowledgements}
We thank Natalie Clarke for their administrative support of this work. Funding: The PINNACLE Consortium is funded by a Wellcome Trust Collaborative Award , “Deciphering AMD by deep phenotyping and machine learning (PINNACLE)”, ref. 210572/Z/18/Z.
}

\clearpage
\bibliographystyle{naturemag}
\bibliography{bibliography}

\clearpage
\setcounter{section}{0}
\renewcommand\thesection{\Alph{section}}

\section{Supplementary material}
\setcounter{secnumdepth}{3}

{

\subsection{Definitions and roles of the junior and senior ophthalmologists}
\label{supplemental:junior_senior_ophthalmologists}
In this study the terms 'junior ophthalmologist' and ‘senior ophthalmologists’ distinguish between clinicians in non-consultant and consultant roles, respectively. This categorization was used to allocate tasks related to data annotation and arbitration. In particular, six junior ophthalmologists provided first- and second-reader labels on the test dataset for the analysis in Sections \ref{sec:results_staging}, \ref{sec:results_referral} and \ref{sec:results_biomarkers}. Two of the junior ophthalmologists curated specialist reports used to generate curriculum part 2 (see Section \ref{sec:method:specialist}). Finally, two senior ophthalmologists arbitrated the first- and second-reader test labels, and assessed the reports written by both the junior ophthalmologists and the VLMs in Figure \ref{fig:figure_4}. 

The reported duration of experience ophthalmologists was counted as time spent practicing in ophthalmological clinics, and did include time spent pursuing their medical degree. All ophthalmologists practice at leading hospitals and ophthalmology institutions in the UK and Austria, including Southampton Eye Unit, Moorfields Eye Hospital and the Medical University of Vienna, and are considered experts in the interpretation of retinal OCT images for AMD.

\subsection{Data curation flowchart}
We summarize the curation and use of all datasets and annotations collected for this study in a flowchart in Figure \ref{fig:data_curation}.

\begin{figure}
    \centering
    \includegraphics[width=0.85\linewidth]{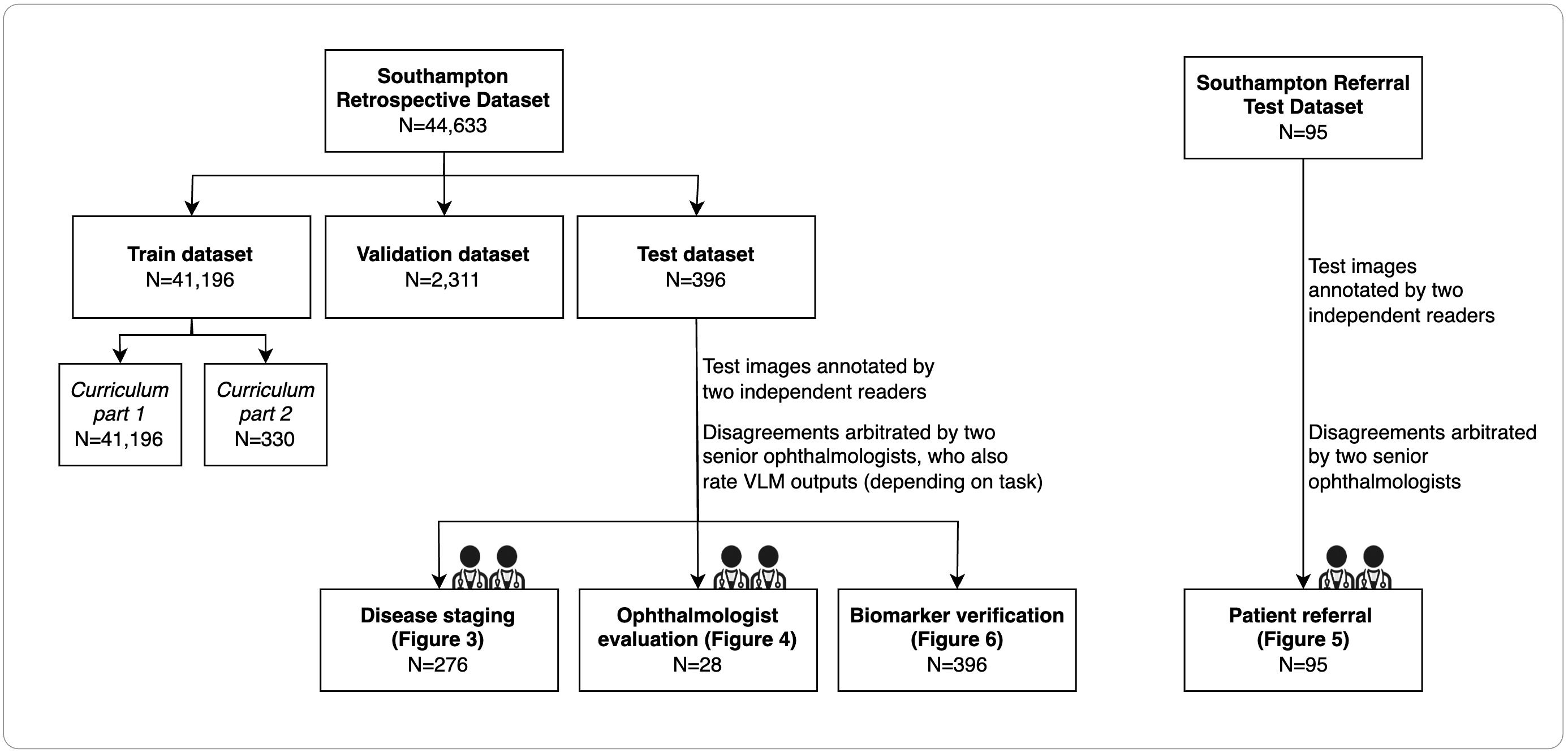}
    \caption{{ Flowchart summarizing the curation of data used to train, validate RetinaVLM models, and test data used to evaluate all VLMs and junior ophthalmologists.}}
    \label{fig:data_curation}
\end{figure}

\subsection{NLP metrics for report generation}

\begin{table}[h]
\centering
\caption{NLP performance metrics for LLaVA-Med and RetinaVLM-Specialist (rounded to two decimal places).}
\hspace{4mm}
\label{table:nlp_metrics}
\begin{tabular}{lccccccc}
\toprule
\textbf{Model} & \textbf{BLEU-1} & \textbf{BLEU-4} & \textbf{METEOR} & \textbf{ROUGE-1} & \textbf{ROUGE-2} & \textbf{ROUGE-L} & \textbf{BERTScore} \\
\midrule
LLaVA-Med & 12.16 & 1.15 & 16.33 & 19.90 & 4.36 & 12.88 & 83.58 \\
RetinaVLM-Specialist & 32.11 & 7.77 & 29.11 & 44.16 & 17.19 & 29.16  & 89.49 \\
\bottomrule
\end{tabular}
\end{table}

In addition to the evaluations by senior ophthalmologists in Figure \ref{fig:figure_4}, we also evaluate reports generated by VLMs using using a series of natural language processing (NLP) metrics. 
We have computed seven metrics including BLEU-1 and BLEU-4 \cite{papineni2002bleu}, METEOR \cite{banerjee2005meteor}, ROUGE-1, ROUGE-2 and ROUGE-L \cite{lin2004rouge}, and BERTSCORE-F1 \cite{zhang2019bertscore}.
Each was used to compare the performance of the Llava-Med foundation model and RetinaVLM-Specialist against the junior ophthalmologists. 

We find that RetinaVLM-Specialist achieved higher scores than LLaVA-Med in every metric (see Table \ref{table:nlp_metrics}), supporting the senior ophthalmologists’ findings of improved performance over Llava-Med. The relative improvement in performance was high in all metrics except for BERTScore, which is known to be insensitive to subtle differences in radiological reports.

\subsection{Ablation experiments on hospital and scanner type}

\begin{figure}
    \centering
    \includegraphics[width=0.85\linewidth]{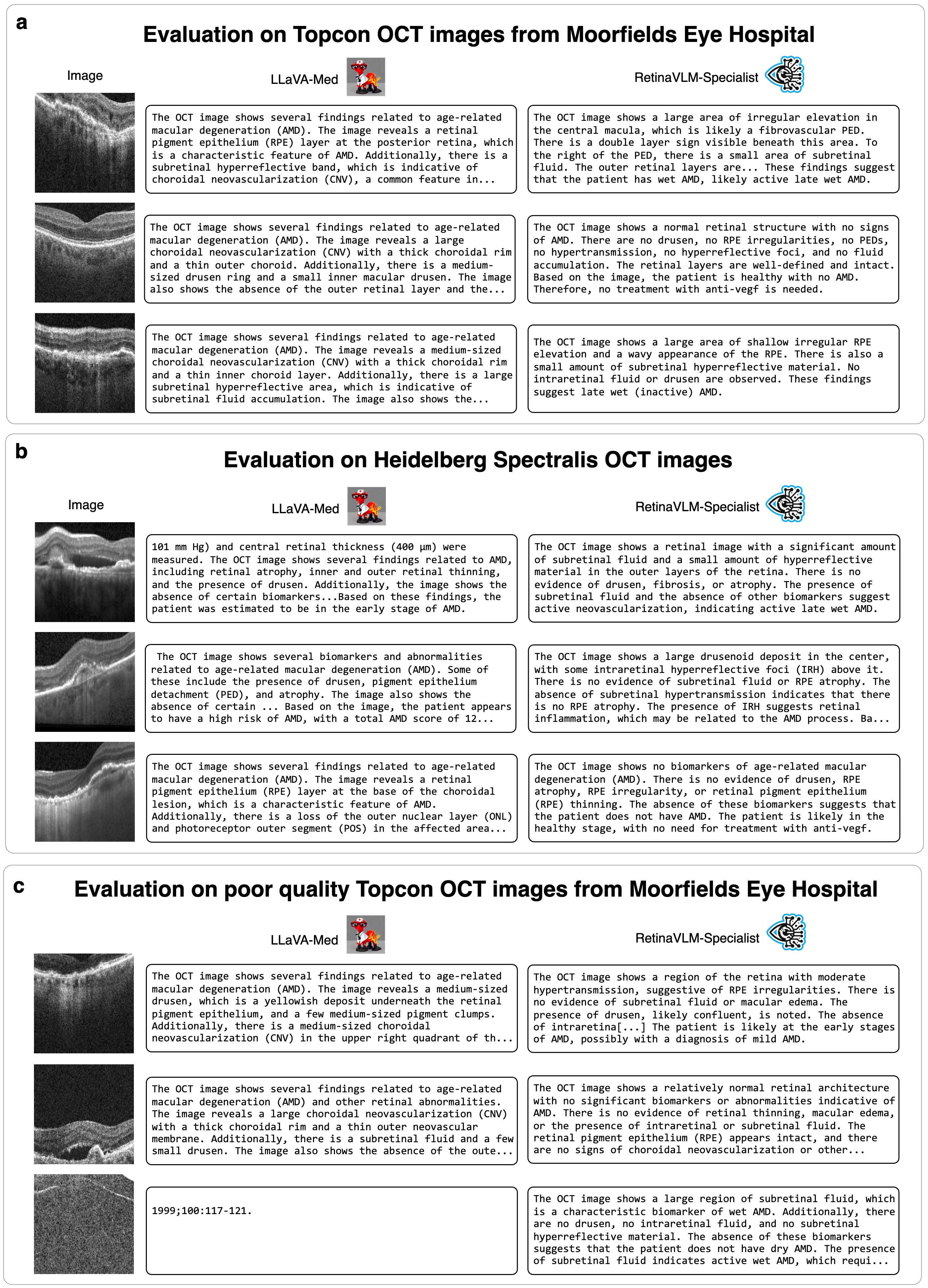}
    \caption{{ RetinaVLM-Specialist and LLaVA-Med performance on external datasets and image types. \textbf{(a)} Testing on Topcon OCT images from Moorfields Eye Hospital showed no observable performance degradation compared to Southampton Eye Unit images. \textbf{(b)} Performance on Heidelberg Spectralis scans, which generally have a higher signal-to-noise ratio, remained comparable to Topcon images, despite the model not being specifically trained on them. \textbf{(c)} Poor image quality from manually selected Topcon OCT images led to consistently inaccurate or hallucinated outputs, demonstrating the need for improved handling of low-quality data in future versions of RetinaVLM.}}
    \label{fig:external_data_tests}
\end{figure}

Our RetinaVLM models were trained on Topcon OCT images from the Southampton Eye Unit. However, in order to begin to understand the limits of these models we qualitatively test them on three small datasets that contain images from different hospitals and scanners. All images used in this experiment were processed using the same method as those in the main study (see Section \ref{sec:method:oct_dataset} for details). To evaluate VLMs on these images we also used the same prompts as in the clinical evaluation study described in Section \ref{sec:method:report_ccc}.

In the first scenario we tested VLMs on Topcon OCT images from another hospital. On images from Moorfields Eye Hospital we did not find any noticeable degradation in performance compared to images from Southampton Eye Unit (see Figure \ref{fig:external_data_tests}a). 

We also test RetinaVLM on scans from the Heidelberg Spectralis scanner. These typically have a higher signal to noise ratio than Topcon scans, though are less widely available. Despite not being trained on Heidelberg Spectralis images, in many cases RetinaVLM-Specialist performs similarly well as on Topcon images (see Figure \ref{fig:external_data_tests} b). We attribute the relatively strong performance to the use of contrastive learning to train the image encoder. Contrastive learning explicitly trains models to become robust to variations in brightness, contrast, and other visual augmentations, which could lead to better generalization across OCT scanners. However, in several cases (such as in the third example) degradations in performance were observed as RetinaVLM failed to detect the presence of clear imaging biomarkers. This indicates that curricula used to train future iterations of RetinaVLM must include a greater variety of retinal imaging modalities.

Finally, we evaluated our model on low-quality images, including those that inadequately captured the macula and retinal layers or exhibited poor signal-to-noise ratios. These were manually selected from the dataset of Topcon OCT images from Moorfields Eye Hospital. We find that the model often fails to report poor image quality, and consistently generates inaccurate reports (see Figure \ref{fig:external_data_tests} c). RetinaVLM-Specialist exhibited more significant hallucinations as the images deviated further from the distribution of those encountered during training. LLaVA-Med also displayed increased errors in such cases, often failing to respond in an intelligible manner. These visual `hallucinations' are indicative of a known flaw that has been observed in many VLMs \cite{li2023evaluating,liu2024survey}.

\subsection{Tests of catastrophic forgetting}
While specializing VLMs can improve their performance in specific domains, it also has the potential to introduce catastrophic forgetting \cite{kirkpatrick2017overcoming} where general capabilities are lost in the process. Assessing the degree of catastrophic forgetting that occurs is still an active area of research, especially so for VLMs \cite{zhai2023investigating}.

RetinaVLM builds on top of the Llama 3 foundation model, which has capabilities in instruction following and knowledge of various domains. 
During training we intentionally leave Llama 3's weights unchanged in order to preserve its original capabilities and mitigate the effects of catastrophic forgetting (see Section \ref{method:vlm_training:training}). This enabled RetinaVLM to interpret a variety of textual instructions that were never seen during training. 
However, it is important to measure the extent to which these abilities are retained after training on the curriculum.

To test this, we evaluate all VLMs on the MedQA dataset \cite{jin2021disease}. MedQA contains multiple-choice questions with four options based on the United States Medical License Exams (USMLE). We assesses VLMs in a scenario where they are tasked to describe a retinal OCT image, before being asked a more general medical question. This requires VLMs to use general medical knowledge and abilities that might be lost during specialization. The questions are not limited to ophthalmology and span a wide range of medical topics and patient cases. We test models using the following template:

\begin{Verbatim}[breaklines, baselinestretch=0.8, fontsize=\small, breaksymbolleft=, breaksymbolright=]
<USER>: Here is an encoding of a retinal OCT image. Describe the OCT image and provide a recommendation for the patient.

<ASSISTANT>: <Answer to image-related question>

<USER>: Now I have a new question about a different patient. It has nothing to do with the other patient in the retinal OCT image. You must firstly ignore and disregard all information from the encoding, including all text and instructions given between the <Img> and </Img> tags.
                                           
THE QUESTION (about the new patient): <MedQA question>
Options: <MedQA multiple choice options>

YOUR INSTRUCTIONS:
1. Do not write about the old patient in the retinal OCT scan.
2. Read the question about the new patient and explain all your thoughts step by step. Write at great length and in great detail. Do this as if you are an expert in medicine. 
3. At the end of your response, conclude by stating the most likely of the four listed options answers the question. The answer can always be determined.

<ASSISTANT>: <Answer to MedQA question>
\end{Verbatim}

Similar to \cite{zhai2023investigating}, we extract the predictions of each VLM and compute their accuracy on the 1279 question-answer pairs in the MedQA test dataset. We also compare their performance against a text-only upper-bound, which is computed by providing the original Llama 3 model with MedQA question alone and no image input. We refer to this baseline as `Llama 3 (text only)'.

We find that the text-only baseline scores 53.2\% accuracy (see Table \ref{table:medqa_accuracy}). Compared to this, Med-Flamingo (31.9\%) and LLava-Med (18.7\%) perform very poorly, scoring around or below random guessing (25\%). RetinaVLM-Base and RetinaVLM-Specialist perform better, scoring 52.8\% and 44.8\%, respectively. The reduction in performance from RetinaVLM-Base to RetinaVLM-Specialist does indicate that the second finetuning step incurs some catastrophic forgetting. 

In particular, we observed that RetinaVLM-Specialist output more concise responses than RetinaVLM-Base. This characteristic did not affect its ability to follow complex instructions related to the interpretation of retinal OCT images in the main study. However, on MedQA it reduced its ability to perform multi-step chain-of-thought reasoning, which is known to improve the performance of models on standardized tests \cite{wei2022chain}. Future future work can address this reduction by increasing the diversity of curriculum part 2 to involve longer, multimodal questions that require multi-step solutions.

\begin{table}[h]
\centering
\caption{{ MedQA accuracy (4 options) for baseline foundation medical VLMs, RetinaVLM models, and a text-only upper bound using Llama 3 8B-Instruct.}}
\hspace{4mm}
\label{table:medqa_accuracy}
\begin{tabular}{l c}
\toprule
\textbf{Model} & \textbf{MedQA Accuracy (4 options)} \\
\midrule
Med-Flamingo & 31.9\% \\
LLaVA-Med & 18.7\% \\
RetinaVLM-Base & 51.3\% \\
RetinaVLM-Specialist & 42.5\% \\
Llama 3 8B-Instruct (text only) & 53.2\% \\
\bottomrule
\end{tabular}
\end{table}

}

{{}
\subsection{Sample image reports from ChatGPT-4o}
In Figure 
\ref{fig:chatgpt_output} We show additional reports written by ChatGPT-4o, including verbose versions of the three discussed in Section \ref{sec:results_ccc_evaluation} and displayed in Figure \ref{fig:figure_4}. Reports were generated using the prompts and methodology described in Section \ref{sec:method:report_ccc}.

\begin{figure}[t]
    \centering
    \includegraphics[width=0.85\linewidth]{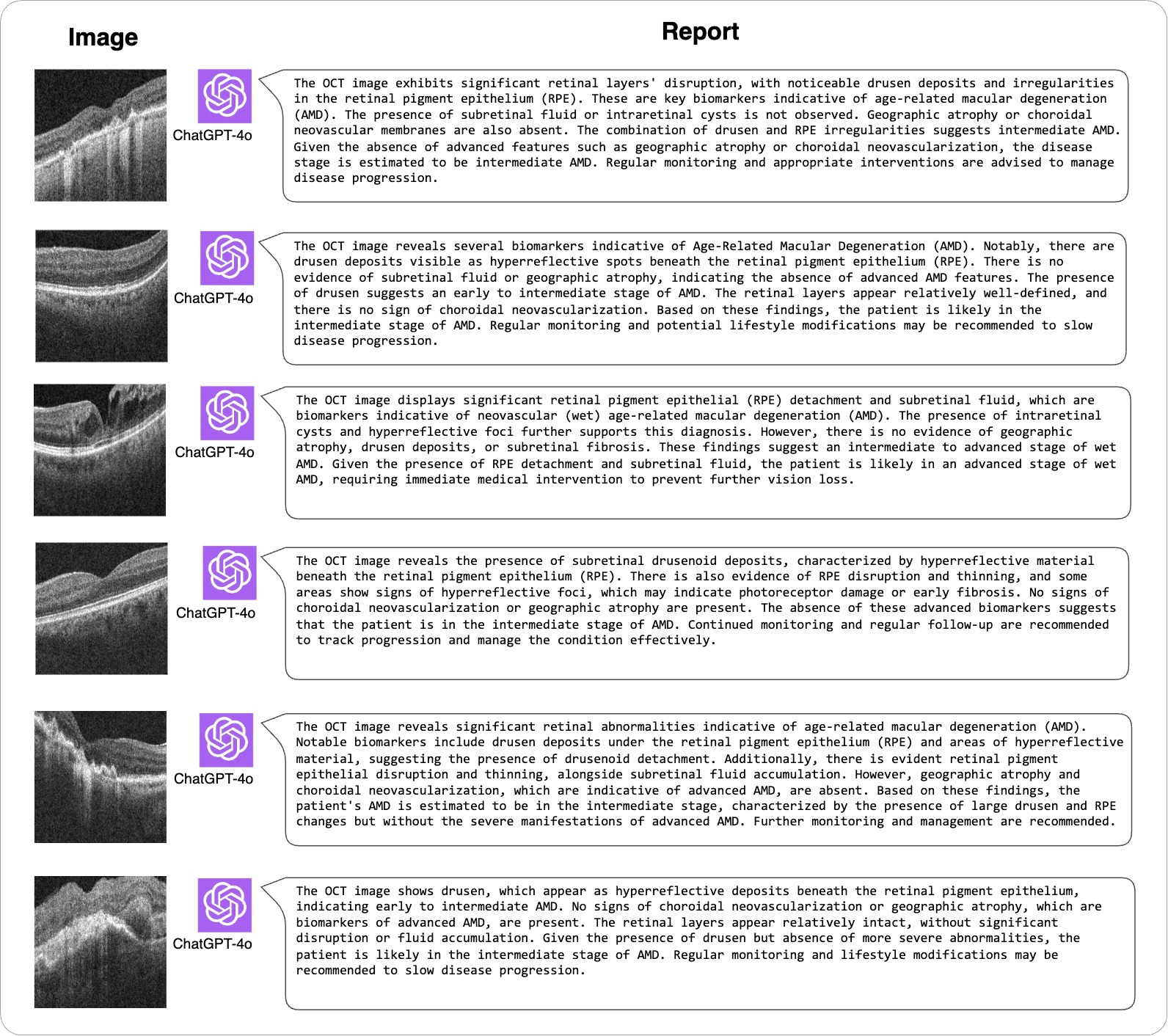}
    \caption{Six sample image reports written by ChatGPT-4o.}
    \label{fig:chatgpt_output}
\end{figure}

}

{{}
\subsection{Effect of staging guidelines on ChatGPT-4o}

\begin{figure}[t]
    \centering
    \includegraphics[width=0.85\linewidth]{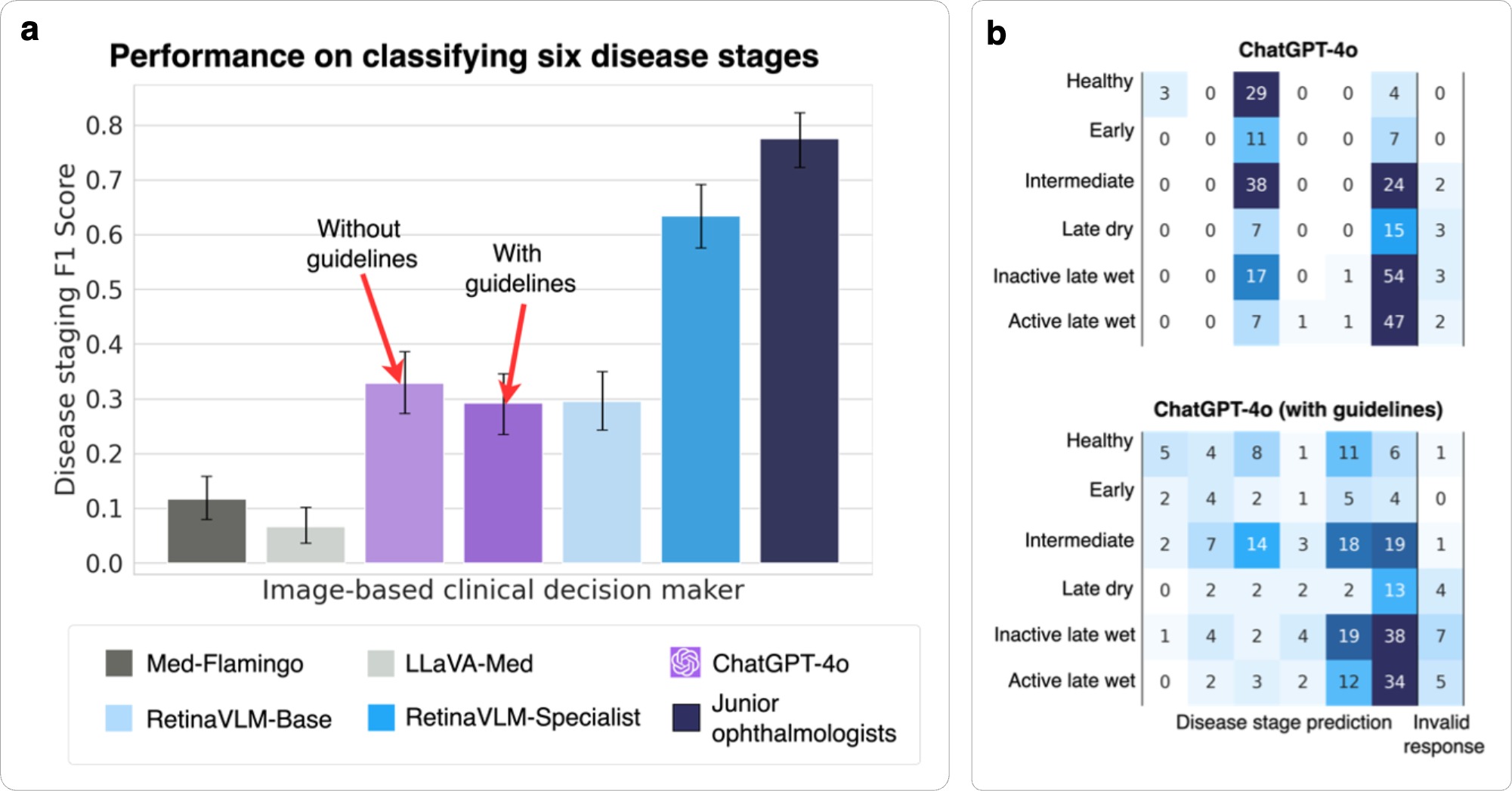}
    \caption{\textbf{(a)} Performance of ChatGPT-4o in disease staging with and without guidelines on how to classify disease in a retinal OCT image. \textbf{(b)}  Effect of including guidelines on staging predictions by ChatGPT-4o.}
    \label{fig:chatgpt_with_guidelines}
\end{figure}

In disease staging GPT-4o performed well in distinguishing non-late from late AMD stages, but did not distinguish active late wet AMD from intermediate, late dry, and inactive late wet AMD compared with RetinaVLM-Specialist (see Figure \ref{fig:figure_3}b, \ref{fig:figure_3}c). Moreover, it did not discern healthy, early and intermediate AMD. To improve the GPT-4o baseline we experimented with providing the disease guidelines that document the six disease stages in finer detail. To this end, we prepended the following text to the prompt given to the model to write disease staging reports (defined in Section \ref{sec:method:disease_staging}). 

\begin{Verbatim}[breaklines, baselinestretch=1.0, fontsize=\fontsize{11}{12}\selectfont, breaksymbolleft=, breaksymbolright=]
Below are guidelines for assessing the AMD stage from a retinal OCT image.

<DiseaseStagingGuidelines>
\end{Verbatim}

Where \texttt{<DiseaseStagingGuidelines>} was replaced with the full staging guidelines documented in verbose form in Figure \ref{extended_fig:curricula}). The results of this experiment are shown in Figure \ref{fig:chatgpt_with_guidelines}. Though this successfully resulted in ChatGPT-4o using all disease stages, it also resulted in a slight drop of performance from 0.33 to 0.29 (see Rebuttal Figure 1). As such, we have elected to report the performance of ChatGPT-4o without guidelines in the main paper. Overall, while ChatGPT-4o is proficient in explaining differences between these disease states in language, our analysis indicates it cannot yet make these distinctions in images.
}

\subsection{Question-answer generation prompts (curriculum part 1: introduction to retina)}
\label{supplemental:curriculum_part1}
The limited scope of the first part of the tabular reports led us to use only one prompt (see Section \ref{supplemental:tabular_qa_prompt}) to create the 408,545 question-answers in curriculum part 1,  `introduction to retina'.

\subsubsection{General question-answering module}
\label{supplemental:tabular_qa_prompt}
  I am constructing a dataset to train a model to answer questions based solely on OCT images. 
  
  The model will access ONLY the image to deduce attributes. For context, the image in question has attributes as follows:
  
  <ReportText>
  
  Based on these attributes, generate a numbered list of diverse questions and answers. 
  
  Ensure the format is:
  
  1. Q: [Question about an image attribute]
  
  A: [Specific answer deduced from the image]
  
  Rules:
  
  - Questions should be crafted in a way that they don't explicitly state the attribute values, but the answers should be based on them. All attributes can be determined from the image.
  
  - Incorporate both yes/no and open-ended styled questions, but always provide a definitive answer in the answer section.
  
  - Occasionally touch on patient outcome/treatment.

\subsection{Question-answer generation prompts (curriculum part 2: advanced retinal specialism)}
\label{supplemental:qa_prompts}

Creating a large quantity and variety of question-answer pairs was important for preventing overfit when finetuning on the 330 images that constitute curriculum part 2.
The prompts each focus on a different aspect of question answering and report writing, but are expected to have some overlap in the question-answer pairs they create. The six modules used to generate up to 230 question-answers per image report, as shown in Figure \ref{fig:figure_2}, are covered by the following 10 prompts.

\begin{enumerate}
    \item Advanced biomarkers (Up to 30 total question-answers)
        \begin{itemize}
            \item Up to 30 question-answer pairs specific to biomarkers (see Section \ref{supplemental:advanced_biomarkers_1})
        \end{itemize}
    \item Disease staging definitions (Up to 50 total question-answers)
        \begin{itemize}
            \item Up to 20 question-answers that use the observation and staging guidelines (see Section \ref{supplemental:staging_definitions_1})
            \item Up to 30 question-answers that train the model to replicate any reasoning linking observations to the disease stage that is present in the report (see Section \ref{supplemental:staging_definitions_2})
        \end{itemize}
    \item Staging reasoning (Up to 60 total question-answers)
        \begin{itemize}
            \item Up to 20 question-answers that use the observation and staging guidelines to train the model to explain the diagnosed disease stage in terms of the visible biomarkers (see Section \ref{supplemental:staging_reasoning_1})
            \item Up to 40 question-answers that focus on training the model in differentiating active from inactive late wet AMD (see Section \ref{supplemental:staging_reasoning_2})
        \end{itemize}
    \item Referral reasoning (Up to 25 total question-answers)
        \begin{itemize}
            \item Up to 25 question-answer pairs that focus on training the model to reason about different levels of urgency in patient referral (see Section \ref{supplemental:referral_reasoning})
        \end{itemize}
    \item General visual question-answering (Up to 30 total question-answers)
        \begin{itemize}
            \item Up to 15 question-answers created directly from the report, but specific to one attribute (see Section \ref{supplemental:specific_report_qa})
            \item Up to 15 question-answers created directly from the report, but including longer and more open-ended questions \ref{supplemental:general_report_qa})
        \end{itemize}
    \item Report writing (Up to 35 total question-answers)
        \begin{itemize}
            \item Up to 15 question-answers that focus on replicating the specialist's report (see Section \ref{supplemental:basic_report_writing})
            \item Up to 20 question-answers that all three observational, disease staging and patient referral guidelines to train the model to create more sophisticated reports \ref{supplemental:advanced_report_writing})
        \end{itemize}
\end{enumerate}

\subsubsection{Advanced biomarkers module}
\label{supplemental:advanced_biomarkers_1}

\begin{Verbatim}[breaklines, baselinestretch=0.8, fontsize=\small, breaksymbolleft=, breaksymbolright=]
  I am constructing a dataset to train a model to answer questions based solely on OCT images.



  Below are guidelines outlining what can appear in an OCT image:

  <ObservationGuidelines>



  However, the image the model is being asked about is characterised by the following description:

  DESCRIPTION OF IMAGE: "<ReportText>"

    

  Task: Write 30 varied questions and answers that ask the model about the image.



  Ensure the format is:

    1. Q: [Question or statement to describe the image]

    A: [Augmented version of actual image description]



  Rules:

  - Ask separately about the presence, amount, location and type of some of the biomarkers in the guidelines

  - Ask about the presence or absence of the biomarkers in the guidelines

  - Rather than saying the image/description does not specify/mention a biomarker, instead say 'the image does not show/display/exhibit evidence of' the biomarker UNLESS its presence is already implied by another present biomarker. The model does not see the above description of the image, it's only given the original image when answering questions.

  - Try not to give away too much information included in the image description in the question text. The model must learn to use the image to determine the answer, and not make educated guesses based on the question alone.

  - The answer must be accurate and reflect the same information in the image description.

  - Write nothing except the questions and answers.



  Tips:

  - Example questions: "Does this image show any subretinal fluid?" "Do you see any intraretinal fluid?" "Is there any hypertransmission? "Does the image show a PED?"

  - Answer style variation: Finally, do not start too many questions with "No, ..." or "Yes, ...". Vary the answer style (the biomarker is 'not present', 'is shown', 'exhibits no', 'does contain' etc...)

  - Positive and negative balance: Try to include an even balance of questions with positive responses (i.e. ask the model about each of the biomarkers that were reported in the description) and negative responses (i.e. that biomarker is not present)

    To do this, if you create a question about a biomarker that isn't in the description, try to create a second, similar question about a biomarker that is observable in the image. 

    In order to make the question set not give too much away about the image, you can make paired questions which have positive and negative responses.

    For example, for an image with a PED but no subretinal fluid, if you ask f.e.

      "Q: Is there a PED? If so, where? A: There is a PED present, it's in the center..." 

    you should also create a question with a negated answer f.e.

      "Q: Is there any subretinal fluid? And in what quantity? A: There is no sign of subretinal fluid in the image...".

    This will help you keep an even balance of positive and negative responses, so that the model cannot guess the answer to the question without considering the image.
\end{Verbatim}

\subsubsection{Disease staging definitions modules}
\label{supplemental:staging_definitions_1}
\begin{Verbatim}[breaklines, baselinestretch=0.8, fontsize=\small, breaksymbolleft=, breaksymbolright=]
  I am constructing a dataset to train a model to answer questions based solely on OCT images.



  Below are guidelines outlining what can appear in an OCT image:

  <ObservationGuidelines>

  <DiseaseStagingGuidelines>



  However, the image the model is being asked about is characterised by the following description:

  DESCRIPTION OF IMAGE: "<ReportText>"

    

  Task: Write 20 varied questions and answers that ask the model about the image.



  Ensure the format is:

    1. Q: [Question or statement to describe the image]

    A: [Augmented version of actual image description]



  Rules:

  - Information about the image should not be in the question.

  - The answer must be accurate and reflect the same information in the image description.

  - Write nothing except the questions and answers.



  Tips:

  - Questions should be specific and ask about certain attributes, or sets of attributes. For example "Q: Is there any subretinal fluid in this image?" or "Q: Is the AMD stage intermediate, or is it more advanced?".

  - Ask separately about the presence, amount, location and type of some of the biomarkers. Try to create an even balance of 'yes and 'no' answers.

  - Rather than saying the image does not directly/explicitly specify/mention the presence of an attribute, instead say 'the image does not exhibit/show/display/evidence' the attribute, UNLESS its presence is already directly implied by the presence of another attribute. The model does not see the above description of the image, it's only given the original image when answering questions.

  - Sometimes the desired output format should be specific in the question (f.e. Answer with 'yes' or 'no', or answer by stating if the image 'does' or 'does not' contain the biomarker in question.)
\end{Verbatim}

\subsubsection{Disease staging from the report module}
\label{supplemental:staging_definitions_2}

\begin{Verbatim}[breaklines, baselinestretch=0.8, fontsize=\small, breaksymbolleft=, breaksymbolright=]
  I am constructing a dataset to train a model to answer questions based solely on OCT images.



  Common disease stages for age-related macular degeneration are: healthy (no-AMD), non-AMD pathology, early AMD, intermediate AMD, late dry AMD, late wet (inactive) AMD, late wet (active) AMD

  These stages may be referred to with slightly different names in the image description.

  If the description doesn't specify whether the late wet AMD is inactive or active, then you shouldn't either (just see how to description refers to it, but don't feel the need to copy it verbatim)



  However, the image the model is being asked about is characterised by the following description:

  DESCRIPTION OF IMAGE: "<ReportText>"

    

  Task: Write 30 varied questions and answers that require the model to estimate the patient's disease stage from the image.



  Ensure the format is enumerated:

    1. Q: [Question or statement to describe the image]

    A: [Augmented version of actual image description]



  Rules:

  - If the estimated disease stage is not provided in the report, do not write ANY questions and answers. Simply write "No disease stage in report".

  - Try not to give away too much information included in the image description in the question text. The model must learn to use the image to determine the answer, and not make educated guesses based on the question alone.

  - The answer must be accurate and reflect the same information in the image description.

  - The questions and answers must vary in their style, formulation and vocabulary.

  - Write nothing except the questions and answers.



  Tips:

  - Most questions should ask the model to first estimate the disease stage

    For example, "Decide the most advanced AMD stage supported by the image, and explain your reasoning by noting any biomarkers most relevant to that stage."

  - If the explanation or reasoning for the disease stage is given in the report, make sure to include that in the model's answers

  - Questions 1 to 20 should first ask the model to decide/estimate/determine/identify/... the (most advanced) disease stage, and then explain their answer.

  - Questions 21 to 30 questions should ask for the biomarkers and then the disease stage, such as "Describe any relevant/notable/significant biomarkers, and link them to the most likely disease stage" (which will be the one in the report)
\end{Verbatim}

\subsubsection{Disease staging reasoning (with guidelines)}
\label{supplemental:staging_reasoning_1}
\begin{Verbatim}[breaklines, baselinestretch=0.8, fontsize=\small, breaksymbolleft=, breaksymbolright=]
  I am constructing a dataset to train a model to answer questions based solely on OCT images.




  Below is a schema outlining what can appear in an OCT image:

  <ObservationGuidelines>

  <DiseaseStagingGuidelines>



  However, the image the model is being asked about is characterised by the following description:

  DESCRIPTION OF IMAGE: "<ReportText>"

    

  Task: Write 20 varied questions and answers that require the model to perform chain-of-thought reasoning.



  Ensure the format is:

    1. Q: [Question or statement to describe the image]

    A: [Augmented version of actual image description]



  Rules:

  - Information about the image listed in the description should not be used in the question, only in the answer.

  - Never use the word 'description' or 'mention' in the answer, the model does not see the description of the image, it only sees the original image itself.

  - The answer must be accurate and reflect the same information in the image description.

  - Write nothing except the questions and answers.




  Tips:

  - Some questions should ask the model to provide a long and detailed answer to questions like 'Describe all the observable biomarkers in the image and then link these to the most likely disease stage'.

  - One or two questions should ask the model to explain/deduce the highest precedent AMD stage (i.e. the overall AMD stage) by describing the biomarker(s) in the image which belong to the most advanced disease stage, and linking them to the relevant disease stage.

  - Some questions should ask the model to summarise any relevant biomarkers and, explaining its reasoning using the guidelines, conclude with the AMD stage. For example, a drusenoid PED suggests intermediate AMD, but if coupled with subretinal fluid the overall AMD stage becomes active late wet due to the fluid.

  - Some questions should ask the model to fully describe and list all its image observations, and then conclude the presence, absence, location or quantity of a specific biomarker (f.e. 'Describe the OCT image in detail and note any abnormalities, and then tell me if the image contains subretinal fluid.')
\end{Verbatim}

\subsubsection{Disease staging reasoning second module}
\label{supplemental:staging_reasoning_2}

\begin{Verbatim}[breaklines, baselinestretch=0.8, fontsize=\small, breaksymbolleft=, breaksymbolright=]
   I am constructing a dataset to train a model to answer questions based solely on OCT images.



  Below are guidelines outlining what can appear in an OCT image:

  <DiseaseStagingGuidelines>



  However, the image the model is being asked about is characterised by the following description:

  DESCRIPTION OF IMAGE: "<ReportText>"

    

  Task: Write 40 varied questions and answers that require the model to estimate the patient's disease stage from the image.



  Ensure the format is enumerated:

    1. Q: [Question or statement to describe the image]

    A: [Augmented version of actual image description]



  Rules that always apply:

  - If the estimated disease stage is not provided in the report, do not write ANY questions and answers. Simply write "No disease stage in report".

  - If reasoning for the disease stage is provided in the report, you MUST include this logic/nuance in the model's answers.

  - You MUST not give away information about the image description in the question text. The model must learn to use the image to determine the answer, and not make educated guesses based on the question alone.

  - The answer must be accurate and reflect the same information in the image description.

  - The questions and answers must vary in their style, formulation and vocabulary.

  - Write nothing except the questions and answers.



  Rules that are specific to differentiating cases of active, vs inactive, late wet AMD:

  - In cases with an active late wet diagnosis, you MUST make it clear that the PRESENCE OF FLUID is what differentiates active from inactive late wet AMD. Make this clear when explaining the reasoning behind active late wet diagnoses.

    So do not imply that f.e. "The disease stage is late wet AMD (active), as indicated by the subretinal/intraretinal fluid, subretinal hyperreflective material, ...."

    Instead, you MUST explain that f.e. "The overall disease stage is late wet AMD, which is active due to the detection/presence of subretinal/intraretinal fluid. Inactive late wet features include fibrovascular PED, ..."

    Or "Late wet AMD best describes the AMD stage. The detection/presence of subretinal/intraretinal fluid means this is active late wet AMD. Other late wet features include fibrovascular PED, ..."

  - Similarly, in cases with an inactive late wet diagnosis, you MUST make it clear in the model's reasoning that the lack of fluid, in combination with the other biomarkers, is what resulted in the inactive late wet diagnosis.

    For example, "The stage is late wet AMD according to the evidence/presence of subretinal hyperreflective material, but it is inactive as there is no detectable fluid of any kind in the image"

  - The exact formulation of this answer must vary according to the question. Do not copy the examples too many times. Add a lot of diversity in the model's responses.



  Tips:

  - Most questions should ask the model to first estimate the disease stage

  - Some questions should first ask the model for its disease stage prediction, and then ask it to explain its reasoning by noting visible biomarkers that relate to that stage

    For example, "Decide the most advanced AMD stage supported by the image, and explain your reasoning by noting any biomarkers most relevant to that stage."

  - The explanation for the disease stage may already be given in the report, but you can also use the guidelines provided to work out which biomarkers resulted in that disease stage.

  - Questions 31 to 40 questions should ask for the biomarkers and then the disease stage, such as "Describe any relevant/notable/significant biomarkers, and link them to the most likely disease stage" (which will be the one in the report)
\end{Verbatim}

\subsubsection{Patient referral reasoning module}
\label{supplemental:referral_reasoning}

\begin{Verbatim}[breaklines, baselinestretch=0.8, fontsize=\small, breaksymbolleft=, breaksymbolright=]
  I am constructing a dataset to train a model to answer questions based solely on OCT images.



  Below are guidelines outlining what can appear in an OCT image:

  <DiseaseStagingGuidelines>



  <PatientReferralGuidelines>



  However, the image the model is being asked about is characterised by the following description:

  DESCRIPTION OF IMAGE: "<ReportText>"



  Task: Write 25 varied questions and answers that require the model to perform explain its reasoning before making conclusions and recommendations.



  Ensure the format is:

    1. Q: [Question or statement to describe the image]

    A: [Augmented version of actual image description]



  Rules:

  - Information about the image should not be in the question.

  - Never use the word 'description' or 'mention' in the answer, the model does not see the description of the image, it only sees the original image itself.

  - The answer must be accurate and reflect the same information in the image description.

  - Write nothing except the questions and answers.

  - The model sees each questions separately so they will not be seen together. 



  Tips:

  - Some questions should ask the model to list any relevant biomarkers and, based on these, recommend if the patient should be referred or not.

  - Many questions should make a series of requests, by asking the model to write a long and detailed answer reporting all the observable biomarkers, linking those to the most likely disease stage and then summarising the report with a patient referral recommendation.

    For example, 'Describe the all biomarkers in the image, and based off of your observations which AMD best describe the patient. Summarise your report with a referral recommendation that follows the treamtent guidelines.'

  - Some questions should ask the model to recommend if the patient does: not need referral, if they need general attention by a specialist, or if they likely need treatment with anti-vegf based off the models observations

  - Some questions should ask the model to explain/deduce/estimate the patient's risk (i.e. the recommended referral action) by first describing the most advanced or concerning biomarker(s) (i.e. the observable biomarker(s) which belong to the most severe disease stage)
\end{Verbatim}

\subsubsection{Specific report-based questions module}
\label{supplemental:specific_report_qa}

\begin{Verbatim}[breaklines, baselinestretch=0.8, fontsize=\small, breaksymbolleft=, breaksymbolright=]
  I am constructing a dataset to train a model to answer questions based solely on OCT images.



  The image in question is characterised by the following description:

  DESCRIPTION OF IMAGE: "<ReportText>"

    

  Task: Write 15 varied questions and answers that ask the model about the image. They should



  Ensure the format is:

    1. Q: [Question or statement to describe the image]

    A: [Augmented version of actual image description]



  Rules:

  - Try not to give away too much information included in the image description in the question text. The model must learn to use the image to determine the answer, and not make educated guesses based on the question alone.

  - The answer must be accurate and reflect the same information in the image description.

  - Write nothing except the questions and answers.



  Tips:

  - Questions should be specific and ask about certain attributes, or sets of attributes. For example "Q: Is there any subretinal fluid in this image?".
\end{Verbatim}

\subsubsection{General report-based questions module}
\label{supplemental:general_report_qa}

\begin{Verbatim}[breaklines, baselinestretch=0.8, fontsize=\small, breaksymbolleft=, breaksymbolright=]
    I am constructing a dataset to train a model to answer questions based solely on OCT images.



  The image in question is characterised by the following description:

  DESCRIPTION OF IMAGE: "<ReportText>"

    

  Task: Write 15 varied questions and answers that ask the model about the image.



  Ensure the format is:

    1. Q: [Question or statement to describe the image]

    A: [Augmented version of actual image description]



  Rules:

  - The answer should be a modified and augmented version of the actual description. 

  - Try not to give away too much information included in the image description in the question text. The model must learn to use the image to determine the answer, and not make educated guesses based on the question alone.

  - The answer must be accurate and contain the same information as the actual description. However, the order of the sentences as they are written must change and randomly vary.

  - Write nothing except the questions and answers.



  Tips:

  - Some question should be general and ask to describe the image.

  - Other questions should be more specific such as "Q: Is there any subretinal fluid in this image?" that have shorter answers.

  - Example questions/statements might be "Describe the OCT image in detail." or "Can you give me a summary of the image?".

\end{Verbatim}

\subsubsection{Basic report writing module}
\label{supplemental:basic_report_writing}

\begin{Verbatim}[breaklines, baselinestretch=0.8, fontsize=\small, breaksymbolleft=, breaksymbolright=]
  I am constructing a dataset to train a model to answer questions based solely on OCT images.



  The image in question is characterised by the following description.

  DESCRIPTION OF IMAGE: "<ReportText>"



  Task: Write 15 questions and answers that ask the model to describe the image in full. The first ten questions should ask to describe the entire image (with jumbled/permuted/randomised answers), while the final five should be more specific or use segments of the description. 

  

  Ensure the format is:

    1. Q: [Question or statement to describe the image]

    A: [Augmented version of actual image description]



  Rules:

  - The answer should be a modified and augmented version of the actual description. 

  - The answer should contain the same information as the actual description but the order of the sentences as they are written must change and randomly vary

  - The question must not contain any information about the image.

  - Write nothing except the questions and answers.



  Tips:

  - Example questions/statements might be "Describe the OCT image in detail." or "Can you give me a summary of the image?" or "Write a report on this image to be given to an optometrist.".
\end{Verbatim}

\subsubsection{Advanced report writing module}
\label{supplemental:advanced_report_writing}

\begin{Verbatim}[breaklines, baselinestretch=0.8, fontsize=\small, breaksymbolleft=, breaksymbolright=]
  I am constructing a dataset to train a model to answer questions based solely on OCT images.



  Below are guidelines outlining what can appear in an OCT image:

  <ObservationGuidelines>

  <DiseaseStagingGuidelines>

  <PatientReferralGuidelines>



  However, the image the model is being asked about is characterised by the following description:

  DESCRIPTION OF IMAGE: "<ReportText>"



  Task: Write 20 requests for reports that ask the model reason all the way from the observable biomarkers, linking these to the most advanced disease stage, and finally to a referral recommendation for the patient.



  Ensure the format is:

    1. Q: [Request to the model]

    A: [Answer or report deduced using actual image description]



  Rules:

  - Never use the word 'description' or 'mention' in the answer, the model does not see the description of the image, it only sees the original image itself.

  - Try not to give away too much information included in the image description in the question text. The model must learn to use the image to determine the answer, and not make educated guesses based on the question alone.

  - Write nothing except the questions and answers.

  - The model sees each questions separately so they will not be seen together. 



  Tips:

  - Example questions/statements might be "Describe the OCT image in detail." or "Can you give me a summary of the image?" or "Write a report on this image to be given to a retinal specialist.".

  - Ask the model to explain why the AMD stage is not more advanced than it is (i.e. the absence of certain late stage biomarkers)

  - Ask the model to explain why the AMD stage is more advanced than an earlier stage (i.e. the presence of certain late stage biomarkers)

  - A few times, request verbose reports like "Write a report that starts by highlighting the most significant and salient biomarkers, and link these to the most probable disease stage. Conclude with a referral recommendation."
\end{Verbatim}

\end{document}